%% file: main.tex
\documentclass[lettersize,journal]{IEEEtran}
\usepackage{amsmath,amsfonts}
\usepackage{algorithmic}
\usepackage{algorithm}
\usepackage{array}
\usepackage[caption=false,font=normalsize,labelfont=sf,textfont=sf]{subfig}
\usepackage{textcomp}
\usepackage{stfloats}
\usepackage{url}
\usepackage{verbatim}
\usepackage{graphicx}
\usepackage{lineno}
\usepackage{cite}
\usepackage{booktabs}
\usepackage{multirow} 
\usepackage{graphicx} 
\usepackage{caption}
\usepackage{xcolor}
\begin{document}

\title{HarmonPaint: Harmonized Training-Free Diffusion Inpainting}

\author{Ying Li, Xinzhe Li, Yong Du, Yangyang Xu, Junyu Dong, Shengfeng He}

\markboth{}
{Shell \MakeLowercase{\textit{et al.}}: A Sample Article Using IEEEtran.cls for IEEE Journals}



\twocolumn[{
\renewcommand\twocolumn[1][]{#1}
\maketitle
\begin{center}
    \captionsetup{type=figure}
    \includegraphics[width=\linewidth]{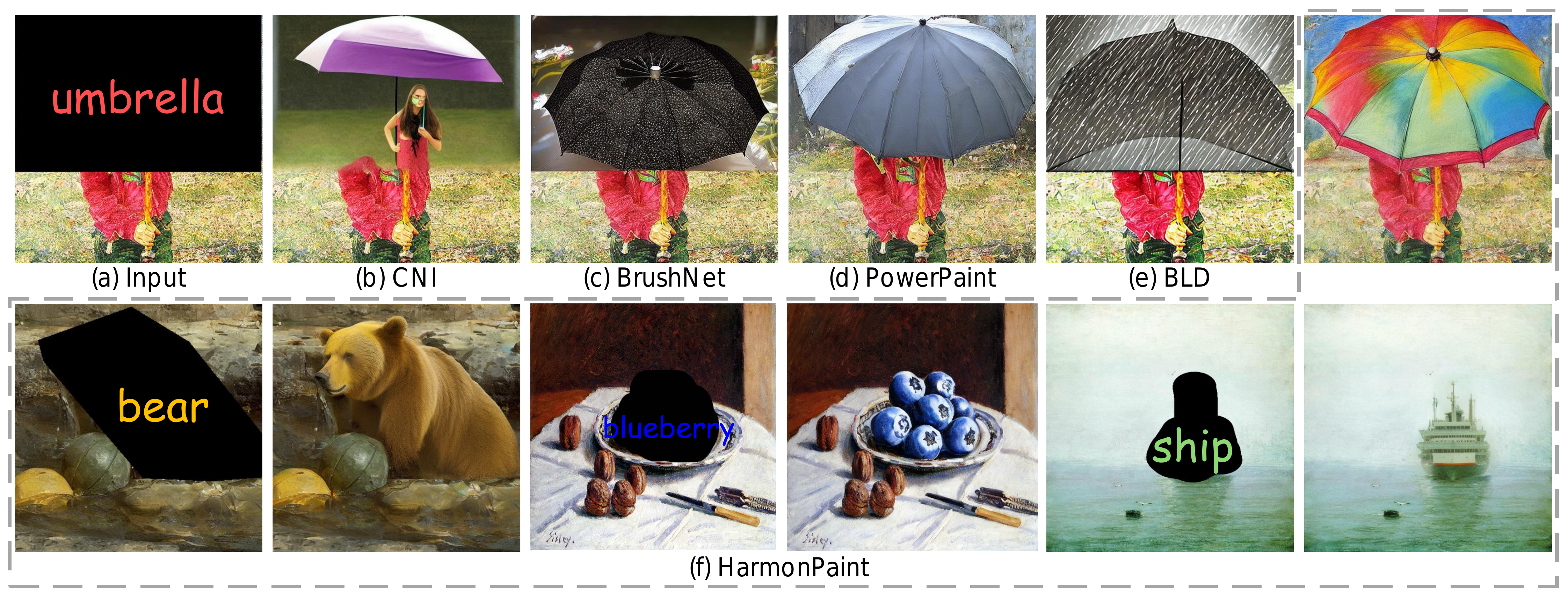}
    \captionof{figure}{We propose \textit{HarmonPaint}, a training-free inpainting framework that achieves harmonized, text-aligned inpainting results. In comparison to existing methods such as (b) ControlNet Inpainting (CNI)~\cite{zhang2023adding}, (c) BrushNet~\cite{ju2024brushnet}, (d) PowerPaint~\cite{zhuang2023task}, and (e) Blended Latent Diffusion (BLD)~\cite{avrahami2023blended}, (f) our approach accurately captures image style and produces structural fidelity results. The bottom row showcases various harmonized inpainting results generated by our method.}
     \label{fig:teaser}
\end{center}
}]

\input{./sec/0_abstract.tex}
\input{./sec/1_introduction.tex}

\input{./sec/2_related}
\input{./sec/3_method}
\input{./sec/4_experiment}

\input{./sec/5_conclusion}

{
\bibliographystyle{IEEEtran}
\bibliography{./egbib}
}

\input{supp}

\vfill

\end{document}

%% file: sec/0_abstract.tex
\begin{abstract}
Existing inpainting methods often require extensive retraining or fine-tuning to integrate new content seamlessly, yet they struggle to maintain coherence in both structure and style between inpainted regions and the surrounding background. Motivated by these limitations, we introduce HarmonPaint, a training-free inpainting framework that seamlessly integrates with the attention mechanisms of diffusion models to achieve high-quality, harmonized image inpainting without any form of training. By leveraging masking strategies within self-attention, HarmonPaint ensures structural fidelity without model retraining or fine-tuning. Additionally, we exploit intrinsic diffusion model properties to transfer style information from unmasked to masked regions, achieving a harmonious integration of styles. Extensive experiments demonstrate the effectiveness of HarmonPaint across diverse scenes and styles, validating its versatility and performance. The source code will be released to the public.


\end{abstract}

%% file: sec/1_introduction.tex
\section{Introduction}
\label{sec:intro}

Diffusion models have recently enabled significant progress in image inpainting, moving beyond traditional texture-based approaches~\cite{ho2020denoising, rombach2022high, saharia2022photorealistic, dhariwal2021diffusion, song2020score}. Unlike previous techniques focused on filling missing regions with generic textures~\cite{barnes2009patchmatch, peng2021generating, zheng2019pluralistic, zheng2022image, zheng2023ciri, fei2023generative}, diffusion models incorporate conditional inputs to produce content-specific inpainting, offering greater flexibility and control over the generated content. This advancement supports a wide range of applications where inpainting can be precisely directed by prompts or contextual information. 

Text-guided image inpainting, which fills masked regions based on textual descriptions, has significant potential in digital art, design, and personalized advertising. Diffusion models have expanded creative possibilities, enabling artists to seamlessly integrate new elements while preserving stylistic integrity. As the demand for stylized inpainting grows, the key challenge is maintaining structural fidelity while allowing for artistic flexibility in partial content regeneration. Despite recent advances, however, current text-guided inpainting methods~\cite{rombach2022high, fei2023generative, lugmayr2022repaint, avrahami2022blended, xie2023smartbrush, zhang2023adding, wang2023imagen} encounter difficulties in producing harmonized inpainting. Frequently, the inpainted content displays unnatural transitions at the edges of masked regions or lacks structural fidelity and stylistic harmony with the rest of the image (see Fig.~\ref{fig:teaser}b). These challenges are exacerbated when inpainting across diverse artistic styles, such as oil paintings or sketches, where focusing on content fidelity alone may disrupt stylistic unity and compromise the visual quality of the overall image.

Recent methods, such as BrushNet~\cite{ju2024brushnet} and PowerPaint~\cite{zhuang2023task}, employ fine-tuning techniques to improve harmony between inpainted regions and surrounding content. However, they often struggle to maintain stylistic harmony across diverse styles due to limited style-specific training data (Fig.~\ref{fig:teaser}d). Similarly, Blended Latent Diffusion~\cite{avrahami2023blended} performs inpainting directly within the latent space, enabling training-free generation. This latent-space blending approach, however, can lead to spatial mismatches between the inpainted content and the prompt due to limited context in the latent space, compromising the harmony of the image (Fig.~\ref{fig:teaser}e). 

In this paper, we introduce \textit{HarmonPaint}, a novel, training-free inpainting framework that embeds inpainting functionality directly into the attention mechanisms of diffusion models. Achieving a seamless integration of inpainted content with the surrounding background, without additional training, presents a significant challenge. HarmonPaint addresses this by adjusting attention processes, enabling diffusion models to generate images aligned with textual prompts while maintaining structural fidelity and stylistic harmony between inpainted regions and the background.

Our approach optimizes inpainting performance through two key objectives: \textit{structural fidelity} and \textit{stylistic harmony}. To achieve structural fidelity, we enhance the attention mechanisms~\cite{vaswani2017attention} within the Stable Diffusion Inpainting model~\cite{rombach2022high}. Unlike previous methods like BrushNet, which simply concatenate the mask and masked image features, we observe that self-attention layers often fail to differentiate between masked and unmasked regions, as they share similar principal components. This blending allows background features to interfere with the inpainting. To address this, we apply a soft mask that reweights the self-attention map between the inpainting and background regions, reducing information crossover so that the principal components of masked regions become distinct from the background. This adjustment enables the diffusion model to clearly identify and refine the inpainting area.

To ensure stylistic harmony, existing methods such as BrushNet and PowerPaint rely on additional module parameters and training, limiting their adaptability beyond specific training data. Instead, we leverage the inherent properties of self-attention: the \( K \) and \( V \) components effectively capture style information. By computing the mean of \( K \) and \( V \) within unmasked regions and propagating them to masked regions, we allow inpainting areas to adopt the overall image style seamlessly, without additional training.

In summary, the contributions of this work are:

\begin{itemize}

\item We introduce a Self-Attention Masking Strategy to control principal components in masked regions, achieving structural fidelity in inpainting.

\item We leverage intrinsic style-capturing properties of diffusion models by propagating style information from unmasked to masked regions for seamless stylistic harmony.

\item Comprehensive qualitative and quantitative experiments validate the effectiveness of HarmonPaint, with ablation studies underscoring the impact of each component.

\end{itemize}

%% file: sec/2_related.tex
\section{Related Work}
\label{sec:Related Work}

\begin{figure*}
	\centering
	\includegraphics[width=0.95\linewidth]{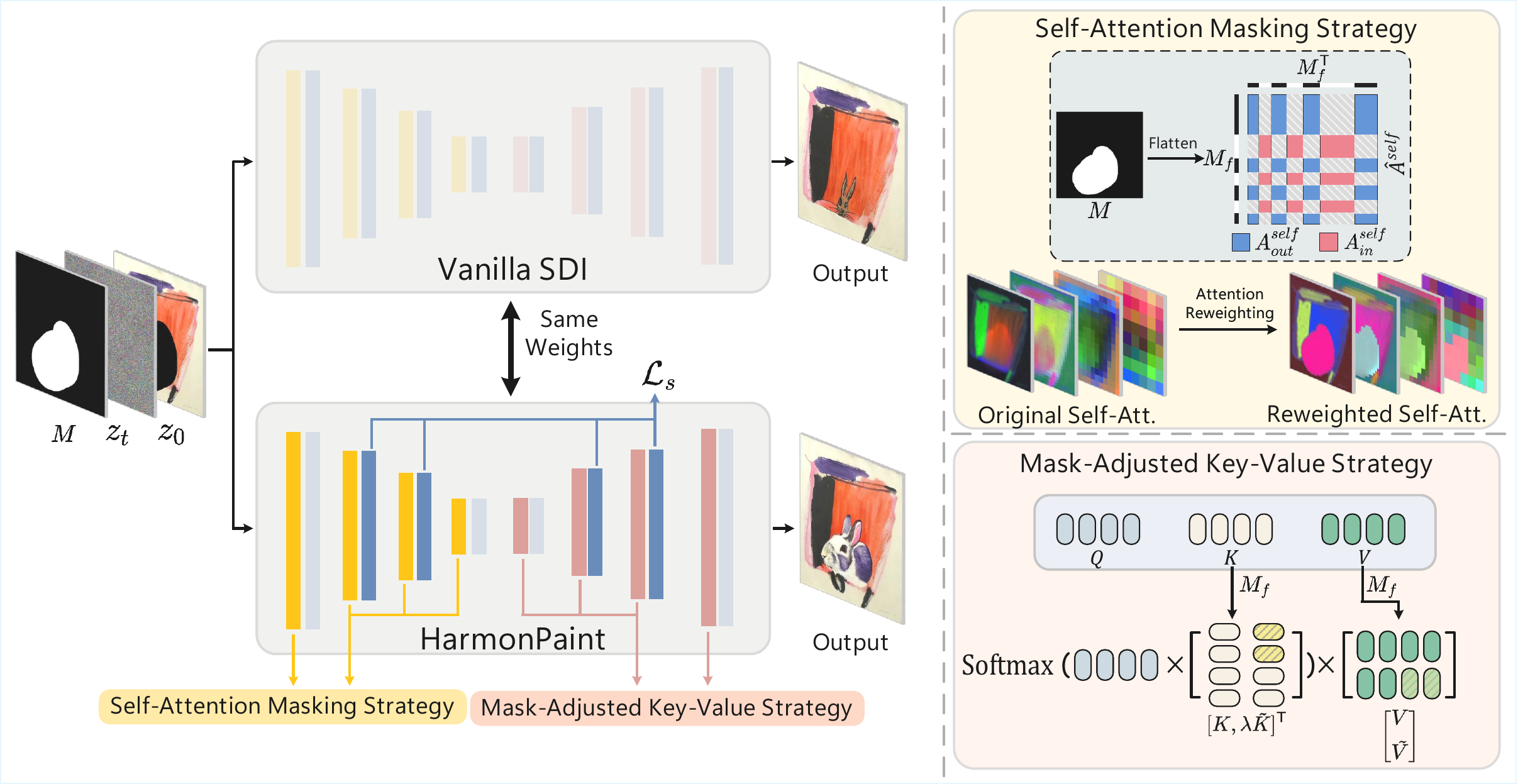}
	\caption{\textbf{Overview of the \textit{HarmonPaint}.} HarmonPaint introduces two key mechanisms to enhance inpainting quality: (1) Self-Attention Masking Strategy, which reweights self-attention maps to ensure structural fidelity, and (2) Mask-Adjusted Key-Value Strategy, transferring style information from unmasked to masked regions to maintain stylistic harmony.}\vspace{-5mm}
	\label{fig:method}
\end{figure*}

\subsection{Text-Guided Image Synthesis}  
Text-guided image synthesis is a generative technique that translates natural language descriptions into images. The introduction of GANs~\cite{goodfellow2020generative, brock2018large, zhang2017stackgan, karras2019style,xu2018attngan,  zhang2021cross,  karras2020analyzing, karras2017progressive, karras2021alias} significantly improved the quality and diversity of generated images, advancing text-to-image synthesis. However, GANs are prone to mode collapse~\cite{srivastava2017veegan} during training. Recently, diffusion models~\cite{ho2020denoising, song2020denoising, ho2022classifier, rombach2022high, saharia2022photorealistic, ramesh2022hierarchical, dhariwal2021diffusion, song2020score} have provided a robust alternative, effectively overcoming training limitations associated with GANs. Diffusion models achieve synthesis by gradually introducing noise through a Markov chain and learning the denoising process. Leveraging large-scale text-image datasets~\cite{schuhmann2022laion}, diffusion-based methods have achieved impressive results. Incorporating attention mechanisms~\cite{vaswani2017attention} for text integration, these methods generate high-quality images that accurately capture semantic information from textual descriptions, facilitating a wide range of downstream applications. However, the high training cost of diffusion models poses a barrier to practical use, especially as many applications require task-specific fine-tuning. Our approach leverages a pre-trained text-to-image diffusion model, exploring its attention modules to create a novel text-guided inpainting framework without the need for training.

\subsection{Image Inpainting}  
Image inpainting seeks to restore or fill missing regions of an image by generating content that blends seamlessly with the original, maintaining visual continuity. Recent diffusion-based inpainting approaches~\cite{fei2023generative, lugmayr2022repaint, avrahami2022blended, avrahami2023blended, xie2023smartbrush, zhuang2023task, chen2025improving, ju2024brushnet, zhang2023adding, wang2023imagen} have greatly improved performance, particularly for text-guided inpainting. Techniques such as Stable Diffusion Inpainting (SDI)~\cite{rombach2022high} and ControlNet Inpainting (CNI)~\cite{zhang2023adding} finetune pre-trained diffusion models by incorporating a mask as an additional input condition. Blended Diffusion~\cite{avrahami2022blended} introduces masks in the noise space, blending the noisy image with CLIP-guided~\cite{radford2021learning} content at each denoising step. However, these methods often compromise visual naturalness and harmony in the inpainted regions. To address this, BrushNet~\cite{ju2024brushnet} incorporates an additional U-Net branch that processes masks in a layer-by-layer approach, mitigating abrupt boundaries in the inpainting results. PowerPaint~\cite{zhuang2023task} uses a dilation operation to avoid overfitting to mask shapes, preserving the overall object structure. UDiffText~\cite{zhao2024udifftext} adopts a method that uses a lightweight character-level text encoder to replace the CLIP module in the stable diffusion model, enabling accurate text generation within the mask. Despite these advances, current methods still lack a coherent spatial layout in their results. Our work introduces a mask-regulated self-attention mechanism that effectively differentiates inpainting areas within the attention map, enhancing structural fidelity.

\subsection{Image Harmonization}  
Image harmonization aims to achieve visual coherence in tasks such as inpainting and compositing, ensuring consistency between foreground and background elements. Recent approaches~\cite{niu2023deep, lu2023tf, zhang2023paste, ren2024relightful, ke2023neural, ling2021region, xue2022dccf, yang2023paint} address layout and style to maintain visual harmony. DCCF~\cite{xue2022dccf} introduces an end-to-end network that learns to apply neural filters for harmonized compositions. Paint by Example (PbE)~\cite{yang2023paint} leverages a pre-trained CLIP model~\cite{radford2021learning} to extract foreground features and injects them into cross-attention layers to improve input image feature alignment. Building on this, PhD~\cite{zhang2023paste} incorporates inpainting and harmonization modules, though it does not address stylistic harmony. TF-ICON~\cite{lu2023tf} proposes noise incorporation and self-attention map adjustments to align style between masked and unmasked regions, though it typically requires a reference image for semantic guidance, limiting its use in text-guided inpainting. 
In contrast, HarmonPaint achieves visual harmony solely from textual prompts without reference images or additional training. Operating in a training-free setting, HarmonPaint adapts to various styles and produces contextually consistent results across diverse visual scenarios.

%% file: sec/3_method.tex
\section{Preliminaries}
\label{section:pre}

Our method builds on the state-of-the-art Stable Diffusion model~\cite{rombach2022high}, utilizing an auto-encoder~\cite{esser2021taming} to efficiently perform the diffusion process within a low-dimensional latent space. Given an image $x_{0}$, a pretrained auto-encoder maps it from pixel space to the latent representation $z_{0} = \mathcal{E}(x_{0})$. During the diffusion process, a Gaussian noise $\epsilon$ is added to $z_{0}$: 
\begin{equation}
	z_{t}=\sqrt{\alpha_{t}} z_{0}+\sqrt{1-\alpha_{t}}\epsilon, \epsilon \sim \mathcal{N}(0, 1),
\end{equation}
where $z_{t}$ denotes the noisy latent at timestep $t$, and $\left \{  \alpha_{t} \right \} $ are the noise schedule. A U-Net architecture~\cite{ronneberger2015u} $\epsilon_{\theta}$ is trained to predict $\epsilon$ for denoising $z_{t}$. Meanwhile, Stable Diffusion encodes the textual prompt via a pre-trained CLIP model~\cite{radford2021learning} to provide an additional input $y$ for conditional guidance. The training objective is as follows:
\begin{equation}
	\mathcal{L}=\mathbb{E}_{z_{t},t,y,\epsilon\sim \mathcal{N}(0, 1)}\left[ {\left \| \epsilon - \epsilon_{\theta}(z_t,t,y)\right \|}_2^2 \right].
\end{equation}

The components most relevant to our work in the U-Net are the self-attention and cross-attention blocks~\cite{vaswani2017attention}. In the self-attention block, the latent image feature $f_{t}$ corresponding to $z_{t}$ are projected to query $Q=W^{q}f_{t}$, key $K=W^{k}f_{t}$, and value $V=W^{v}f_{t}$, where $W^q$, $W^k$, and $W^v$ are learnable matrices. The output of the block ${f}'_{t}$ is computed through a softmax operation as follows: 
\begin{equation}
	{f}'_{t}=A^{self} V, A^{self}=\text{Softmax}(\frac{QK^{\top}}{\sqrt{d}} ).
\end{equation}
Where $d$ is the output dimension of the key and query, and $A^{self}$ represents the self-attention map. Cross-attention block operates similarly but with key and value computed from the token embeddings of the textual prompt.

\section{HarmonPaint}
\label{sec:method}
Achieving harmonized inpainting requires addressing two essential aspects: structural fidelity and stylistic harmony. Structural fidelity ensures that the inpainted content maintains reasonable shape, size, and alignment with the spatial structure and perspective of the image. Stylistic harmony ensures that the inpainted content aligns with the image's original style, avoiding visual dissonance. In the following sections, we introduce novel mechanisms to address each of these challenges.

\subsection{Structural Fidelity}\label{section:contri1}
Research~\cite{cao2023masactrl, tumanyan2023plug} has shown that self-attention maps in diffusion models capture the overall layout of an image. We conduct experiments to further illustrate the role of self-attention blocks in image inpainting, an aspect often overlooked by existing methods. Given an incomplete image, a binary mask (with values of 1 in the masked region) to indicate missing areas, and a textual prompt, we apply Stable Diffusion Inpainting (SDI)~\cite{rombach2022high} for inpainting, with results displayed in Fig.~\ref{fig:method}. It can be observed that the ``rabbit'' generated by SDI has an incomplete body and does not align with the geometric relationships in the surrounding context.

To analyze the cause of this issue, we extract self-attention maps from the U-Net and perform Principal Component Analysis (PCA)~\cite{hotelling1933analysis}. Since skip connections inherently couple encoder and decoder representations, making feature disentanglement challenging, we specifically analyze the encoder, where information propagation in U-Net originates. Fig.~\ref{fig:method} illustrates the three principal components of these self-attention maps, where semantically similar regions share similar colors. We observe that SDI groups the masked and unmasked regions into the same principal component within the encoder of the U-Net, which is suboptimal. Prior research~\cite{li2023faster} suggests that the U-Net encoder extracts latent features from noisy latent inputs, while the decoder uses these features to predict the spatial layout of the image. When the encoder treats masked and unmasked regions as a highly similar component, it struggles to capture the spatial relationships between the inpainted content and surrounding areas, leading to an unrealistic layout.

To address this issue, we propose a Self-Attention Masking Strategy (SAMS). Previous studies~\cite{dosovitskiy2020image} have shown that self-attention mechanisms effectively capture relationships between image patches. Building on this foundation, we partition the self-attention map into three distinct regions based on the inpainting task: interactions within the masked region (object-object, denoted as `obj-obj'), interactions within the unmasked region (background-background, denoted as `bg-bg'), and interactions between the masked and unmasked regions (object-background, denoted as `obj-bg'). By selectively masking the obj-bg interactions, our approach prevents background information from leaking into the object region, leading to a more focused and precise inpainting process. This strategy minimizes unwanted background interference, thereby enhancing the accuracy and quality of the inpainted content.

We first resize the given binary mask to match the dimensions of self-attention map $A^{self}\in \mathbb{R}^{HW \times HW}$ in each self-attention block, resulting in $M\in \mathbb{R}^{H \times W}$. We then flatten $M$ into a one-dimensional vector $M_{f} \in \mathbb{R}^{HW \times 1}$, and incorporate it into $A^{self}$ as follows:
\begin{equation}
	A^{self}_{in}=M_{f} \times M_{f}^{\top} \odot A^{self}, 
\label{eq:self_in}
\end{equation}
\begin{equation}
	A^{self}_{out}=(\mathbf{1}-M_{f}) \times (\mathbf{1}-M_{f})^{\top} \odot A^{self}, 
\label{eq:self_out}
\end{equation}
\begin{equation}
	\widehat{A}^{self} = A^{self}_{in} + A^{self}_{out}, 
\end{equation}
where $A^{self}_{in}$ represents the obj-obj interaction, and $A^{self}_{out}$ represents the bg-bg interaction, corresponding to the pink and blue areas in Fig.~\ref{fig:method}, respectively. During the encoding phase, we replace the original self-attention map ${A}^{self}$ with our modified $\widehat{A}^{self}$ by selectively masking the obj-bg interactions, (represented by the gray areas in Fig.~\ref{fig:method}). This adjustment allows latent features within the masked region to focus more on the inpainting area while reducing dependence on the unmasked region. As shown in Fig.~\ref{fig:method}, features in the masked region have different principal components from those in the unmasked region. Unlike previous approaches~\cite{kim2023dense,dahary2024yourself,sun2024attentive}, which simply differentiate self-attention components inside and outside the mask, our method introduces a more refined segmentation of the self-attention map, tailored to the specific requirements of the inpainting task. This refinement allows SDI to better distinguish masked and unmasked regions, improving structural fidelity.

However, introducing the binary mask directly into the self-attention block may overly hinder information exchange between the masked and unmasked regions. Therefore, we employ a soft mask~\cite{szegedy2016rethinking} in practice: 
\begin{equation}
	\widehat{M}_{f} = (1-\tau)M_{f} + \frac{\tau}{HW}.
\end{equation}
Here, $\tau$ is a smoothing factor, and we replace $M_{f}$ with $\widehat{M}_{f}$ in Eq.~(\ref{eq:self_in}) and Eq.~(\ref{eq:self_out}). 

The structural fidelity of image inpainting depends not only on the relationship between object and background patches but also on the alignment between the inpainted object and the text prompt. To address this challenge, we draw inspiration from~\cite{sueyoshi2024predicated}, which computes the logical relationship between textual descriptions and images. To better adapt this concept to inpainting, we propose Attention Steer Loss, an extension of this approach that specifically focuses prompt-related tokens within the masked region. The key steps are listed below.

At each timestep $t$, we perform a denoising process to extract cross-attention maps at resolutions 16 and 32, as these resolutions have been shown to capture the most semantic information~\cite{hertz2022prompt}. We then compute the averaged cross-attention map $A^{cross} \in \mathbb{R}^{HW \times L}$ at each resolution, where $L$ represents the number of tokens in the textual prompt. For the $i$-th token, we incorporate $M_{f}$ into its attention map $A^{cross}_i \in \mathbb{R}^{HW \times 1}$ as follows:
\begin{equation}
	\widehat{A}^{cross}_i=A^{cross}_i \odot M_{f}. 
\end{equation}
Subsequently, we enumerate the patches of $\widehat{A}^{cross}_{i}$ and define the following loss objective:
\begin{equation}
	\mathcal{L}_{s}=-\sum_{i} \log(1-\prod_{j} (\mathbf{1}-\widehat{A}^{cross}_{i,j})), 
\end{equation}
where $\widehat{A}^{cross}_{i,j}$ denotes the attention of the $i$-th token to the $j$-th patch of the noisy latent. Since $M_{f}$ sets the values in the unmasked region to zero, minimizing $\mathcal{L}_{s}$ encourages the attention of tokens to be fully concentrated within the masked region.

\subsection{Stylistic Harmony}\label{section:contri2}
We frame stylistic harmony enhancement as a localized style transfer task. Traditional style transfer methods, as documented in prior works~\cite{hertz2024style,mu2024fontstudio}, extract attributes such as color and texture from a reference image and apply them to a target content image. In our inpainting task, this paradigm manifests differently: the unmasked region serves as the style reference, while the masked region acts as the content reference. However, unlike conventional style transfer techniques~\cite{hertz2024style,mu2024fontstudio}, which transfer style between separate images, our method performs local style transfer within a single image. This requires extracting style information from specific regions while preserving both style and content consistency across the entire image.

$K$ and $V$ of self-attention in the U-Net decoder is capable of extracting style information~\cite{chung2024style}, and we propose a Mask-Adjusted Key-Value Strategy (MAKVS) to integrate these features into the masked region for stylistic harmony. Taking $K$ as an example, we update $K$ through $M_{f}$ as follows: 
\begin{equation}
\widetilde{K}(i) = \begin{cases}
 K(i),&\text{ if } M_{f}(i)=0, \\
\bar{K}, &\text{+otherwise},
\end{cases}
\end{equation}
where $K(i)$ denotes the features in the $i$-th patch of the image, and $\bar{K}$ is the mean of $K(i)$ over the unmasked region. We consider $\bar{K}$ contains the overall style features of the incomplete image and use it to replace the original style features in the masked region, thereby aligning the style of the masked region with that of the unmasked region. 

\begin{figure}
	\centering
	\includegraphics[width=\linewidth]{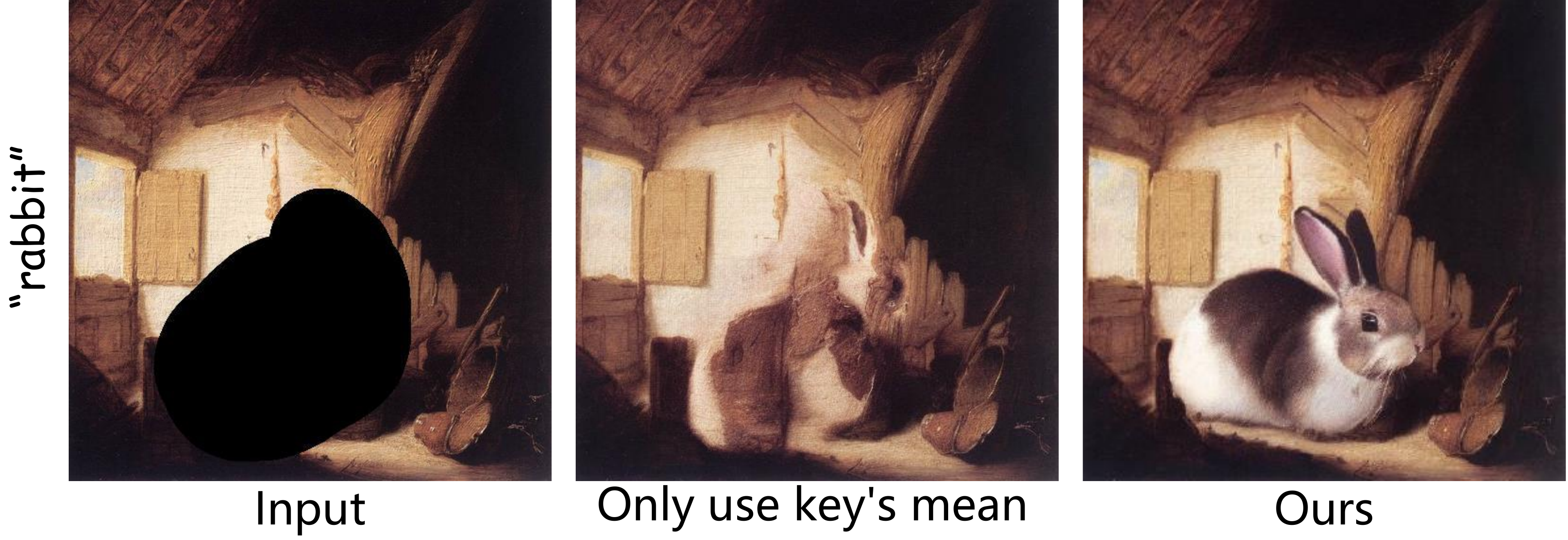}
	\caption{Comparison results between HarmonPaint and the variant with mean-based key aggregation.}
	\label{fig:style consistency}
\end{figure}

We apply a similar operation to obtain $\widetilde{V}$, and then compute the self-attention map $\widetilde{A}^{self}$ and latent feature $\widetilde{f}'_{t}$ as:
\begin{equation}
	\widetilde{A}^{self}=\text{Softmax}(\frac{Q \times [K, \lambda \widetilde{K}]^{\top} }{\sqrt{d}}),
 \label{eq:tilde-self-att}
\end{equation}
\begin{equation}
	\widetilde{f}'_{t} = \widetilde{A}^{self} \times \begin{bmatrix} V \\  \widetilde{V} \end{bmatrix}, 
\end{equation}
where $\lambda$ is a hyperparameter used to control the strength of style transfer. There is a relatively naive approach that we can compute the self-attention map by using $\bar{K}$ directly as: 
\begin{equation}
	\widetilde{A}^{self}_{naive}=\text{Softmax}(\frac{Q \times \widetilde{K}^{\top} }{\sqrt{d}}).
\end{equation}
However, as shown in Fig.~\ref{fig:style consistency}, comparing our approach with a variant that uses only the mean of the key values reveals a critical limitation. While this variant effectively captures the style of the unmasked region, relying solely on it can disrupt content generation. Specifically, using only the key’s mean weakens structural preservation, leading to distortions in the intended object shape, such as the deformation of the 'rabbit' structure. As discussed in Sec.~\ref{section:contri1}, $A^{self}$ encodes the spatial layout of the image. Therefore, in Eq.~(\ref{eq:tilde-self-att}), we introduce $\lambda$ to balance style and structural integrity by concatenating $K$ and $\widetilde{K}$.

\subsection{Efficient Division Strategy}
Current research~\cite{balaji2022ediff, zhang2023prospect} indicates that diffusion models initially generate coarse outlines and shapes of objects, progressively refining details (\textit{e.g.}, style) as the denoising process continues. Accordingly, our method leverages this characteristic to divide our denoising process into two stages for improved performance. We introduce a hyperparameter $\eta \in (0,1)$ to split the total timestep $[0,T]$ into $[0,\eta T]$ and $[\eta T,T]$, and then we focus on enhancing the structural fidelity of the inpainted content in $[\eta T,T]$, while emphasizing stylistic harmony between the masked and unmasked regions in $[0,\eta T]$. This design significantly reduces the computational cost during inference. 

%% file: sec/4_experiment.tex
\section{Experiments}
\label{sec:exp}

\begin{figure*}
	\centering
	\includegraphics[width=\linewidth]{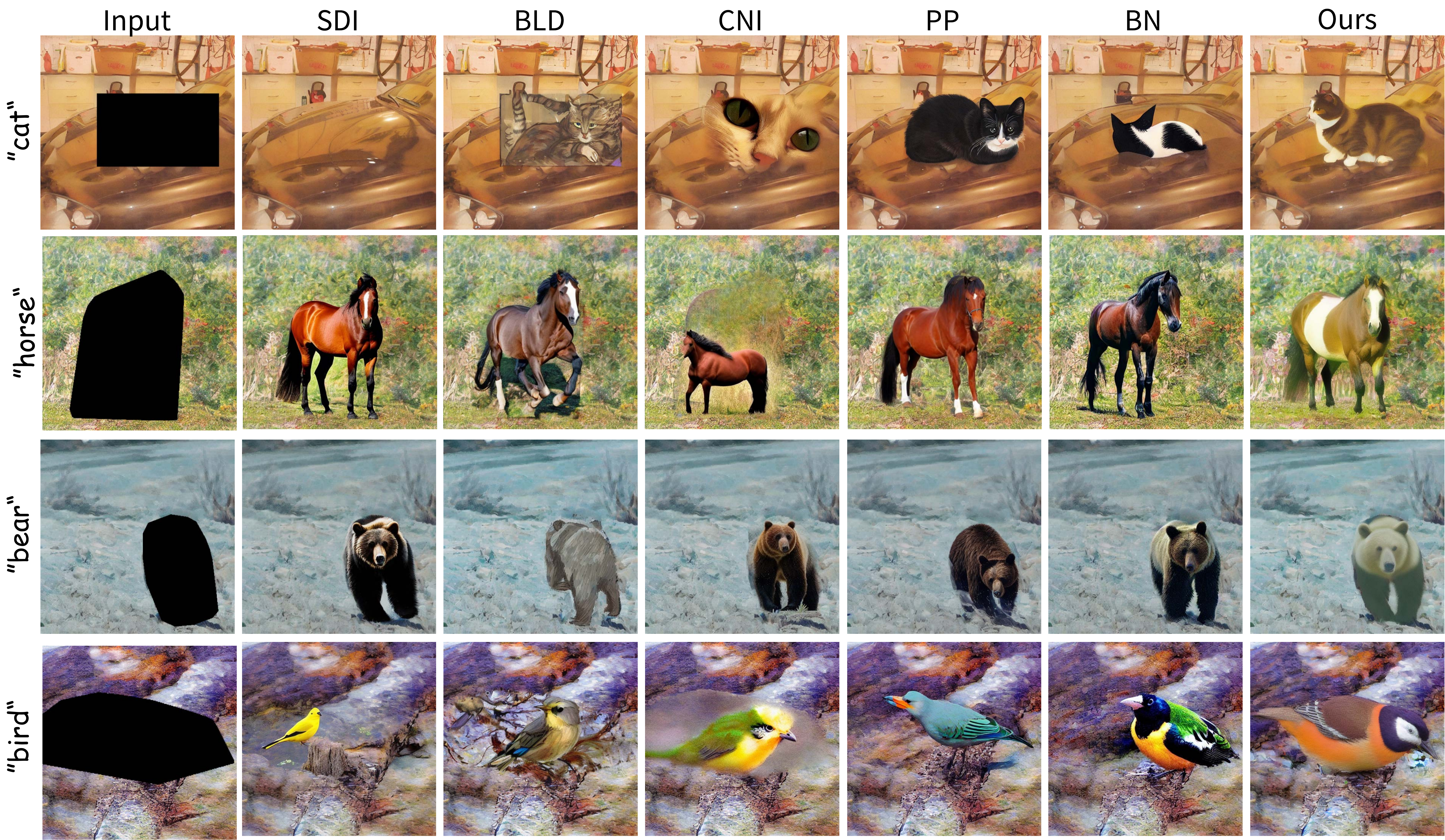}
	\caption{Qualitative comparison with state-of-the-art methods on the Stylized-COCO Dataset.}\vspace{-4mm}
	\label{fig:mscoco}
\end{figure*}

\subsection{Experimental Settings}
\textit{1) Implementation Details:}  
Our method is implemented on top of SDI and operates without additional training. We set the classifier-free guidance scale~\cite{ho2022classifier} to 7.5 and use 50-step DDIM~\cite{song2020denoising} sampling. For the division strategy, we set parameters as follows: $\tau=0.1$, $\lambda=1.4$, and $\eta=0.6$. Self-Attention Masking Strategy (SAMS): is applied to layers 2-6 of the U-Net, while the Mask-Adjusted Key-Value Strategy (MAKVS) is employed in the final 8 layers.

\textit{2) Competitors:}  
We compare our method with several state-of-the-art text-guided inpainting methods, including Blended Latent Diffusion (BLD)~\cite{avrahami2023blended}, ControlNet Inpainting (CNI)~\cite{zhang2023adding}, Stable Diffusion Inpainting (SDI)~\cite{rombach2022high}, BrushNet (BN)~\cite{ju2024brushnet}, and PowerPaint (PP)~\cite{zhuang2023task}. For CNI, BN, and PP, we use their official pretrained models. All competitors are evaluated with their default settings.

\textit{3) Test Benchmarks:}  
Commonly used datasets for image inpainting, such as MSCOCO~\cite{lin2014microsoft}, OpenImages~\cite{kuznetsova2020open}, and ImageNet~\cite{deng2009imagenet}, are primarily designed for natural images and do not suit our aim of testing harmonized inpainting across diverse styles. Therefore, we created a Stylized-COCO dataset by selecting 50 reference images in various styles from WikiArt~\cite{tan2018improved} and applying StyleID~\cite{chung2024style} to transform MSCOCO images into different artistic styles. The dataset includes both segmentation masks and bounding box masks, with MSCOCO captions as text conditions. Additionally, we generated a Stylized-OpenImages dataset following the same procedure, using OpenImages as the base.

\textit{4) Evaluation Metrics:}  
Traditional inpainting metrics, such as FID~\cite{heusel2017gans} and KID~\cite{binkowski2018demystifying}, fail to adequately capture the diversity and style harmonization required in our setting~\cite{jayasumana2024rethinking}. Instead, we evaluate performance using CLIP Score (CS)~\cite{hessel2021clipscore}, Image Reward (IR)~\cite{xu2024imagereward}, Aesthetic Score (AS)~\cite{schuhmann2022laion}, and CLIP Maximum Mean Discrepancy (CMMD)~\cite{jayasumana2024rethinking}. CS measures the alignment between inpainted content and the provided text prompt. IR, a human preference-based model designed for text-to-image tasks, assesses structural fidelity. AS quantifies stylistic harmony using a linear model trained on real image-quality scores. CMMD evaluates distributional differences between the reference and generated image sets by computing the squared MMD distance between their CLIP embeddings.

\begin{figure*}
	\centering
	\includegraphics[width=\linewidth]{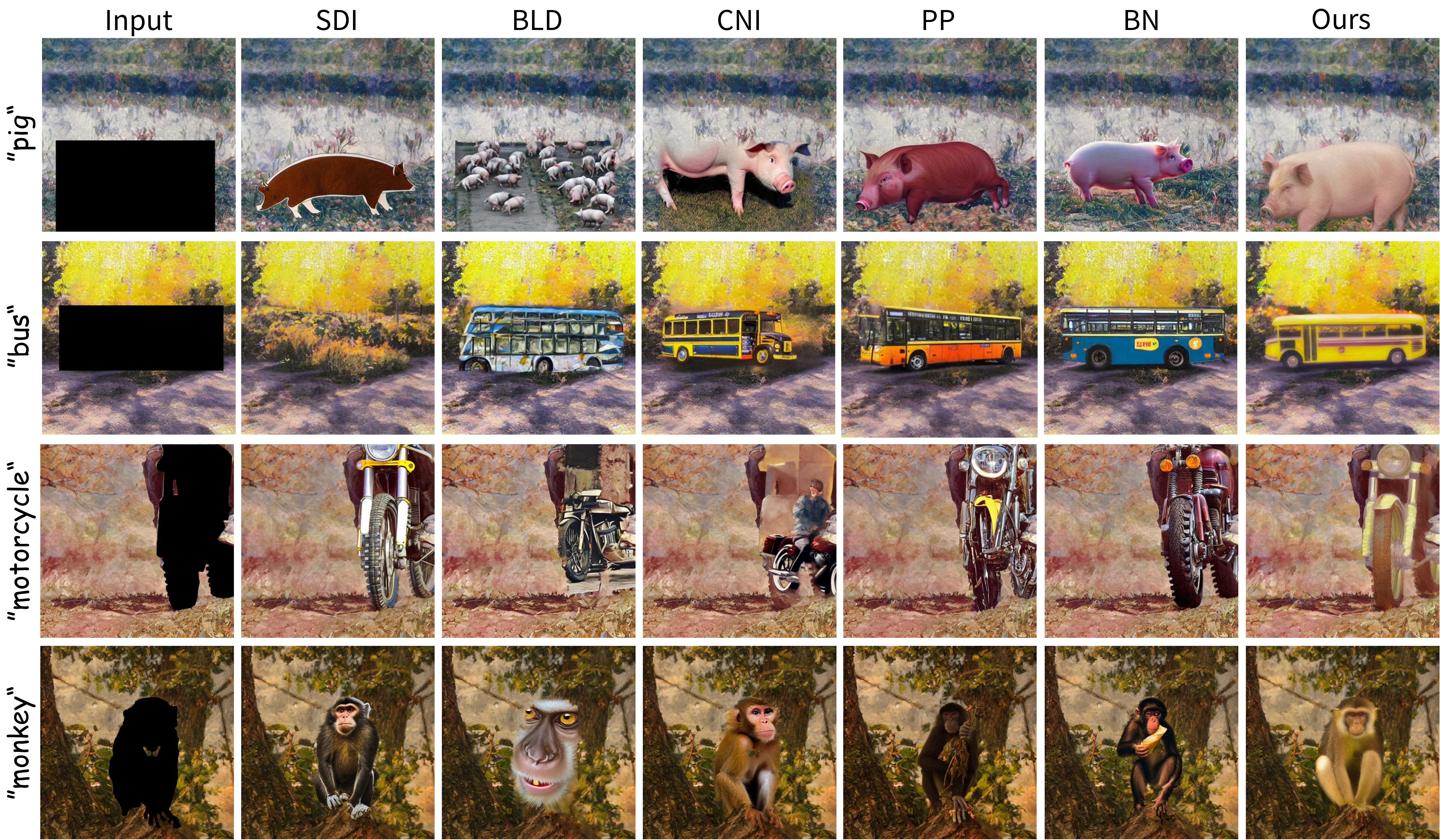}
	\caption{Qualitative comparison with state-of-the-art methods on the Stylized-OpenImages Dataset.}\vspace{-4mm}
	\label{fig:openimages}
\end{figure*}

\subsection{Comparison with State-of-the-Art Methods}
\textit{1) Qualitative Evaluation:}  
Our qualitative results are presented in Fig.~\ref{fig:mscoco} and Fig.~\ref{fig:openimages}. On the Stylized-COCO dataset, we observe that BLD and CNI often produce spatially inconsistent results. For instance, in the first row of Fig.~\ref{fig:mscoco}, while their inpainted content matches the given description, the shape and contour of the ``cat'' fail to harmonize with the surrounding context. This inconsistency arises because these methods overlook the role of self-attention during denoising, where masked and unmasked regions are treated as part of the same principal components in self-attention maps, limiting the diffusion model’s ability to capture spatial relationships. In contrast, our method introduces a SAMS in the U-Net encoder, effectively addressing this limitation and enhancing structural fidelity.  

Competitors also face challenges with stylistic harmony. For example, in the second row of Fig.~\ref{fig:mscoco}, BLD and CNI generate a real-world horse within the masked region, but the style of the inpainted area does not match the unmasked surroundings, resulting in a visually discordant composition. In contrast, our method aligns with the prompt while maintaining stylistic harmony between the masked and unmasked regions. This is achieved through our MAKVS, which extracts style information from the unmasked regions and integrates it into the masked areas.  
On the Stylized-OpenImages dataset, our method similarly achieves harmonized results, demonstrating its effectiveness across diverse styles and scenarios.

\begin{table}[t]
	\renewcommand\arraystretch{1.2}
	\centering
	\footnotesize
        \setlength\tabcolsep{1mm}	
    {
		\begin{tabular}{l||c|c|c|c|c|c|c|c}
			\toprule
			\multirow{3}*{Method}  & \multicolumn{8}{c}{Stylized-MSCOCO}  \\ 
			\cmidrule(lr){2-9}
            & \multicolumn{4}{c|}{Segmentation Mask}& \multicolumn{4}{c}{Bounding Box Mask} \\
            \cmidrule(lr){2-5}
			\cmidrule(lr){6-9}
			&CS$\uparrow$ & IR$\uparrow$ &AS$\uparrow$ &CMMD$\downarrow$ &CS$\uparrow$ & IR$\uparrow$ &AS$\uparrow$ &CMMD$\downarrow$\\	
			\midrule
			BLD & 26.47 & -0.72 & 6.33 & 0.127 & 26.21 & -1.11 & \underline{5.68} &0.093 \\
            CNI & 27.11 & -0.62 & 5.46 & 0.308 &27.55 & -0.89 & 4.99 & 0.207\\
            SDI & 26.70 & \underline{-0.61} & \underline{6.52} & 0.137 & 26.23 & -0.88 & 5.40 & 0.103 \\ 
            BN & 26.88 & -0.63 & 6.25 & \underline{0.109} & 26.13 & -0.99 & 5.00 & \underline{0.090}\\
            PP & \underline{ 28.12} & -0.72 & 6.34 & 0.112& \underline{27.69} & \underline{-0.69} & 5.25 & 0.097\\	
			\midrule
			\textbf{Ours} & \textbf{28.86} & \textbf{-0.56} & \textbf{6.55} & \textbf{0.103} & \textbf{28.27} & \textbf{-0.46} & \textbf{5.96} & \textbf{0.080}\\
			\bottomrule
		\end{tabular}}
	\caption{Quantitative evaluation on Stylized-MSCOCO dataset. }\vspace{-4mm}
	\label{tab:Quantitative_mscoco}
\end{table}

\begin{table}[t]
	\renewcommand\arraystretch{1.2}
	\centering
	\footnotesize
        \setlength\tabcolsep{1mm}	
		\begin{tabular}{l||c|c|c|c|c|c|c|c}
			\toprule
			\multirow{3}*{Method}  & \multicolumn{8}{c}{Stylized-OpenImages}  \\ 
			\cmidrule(lr){2-9}
            & \multicolumn{4}{c|}{Segmentation Mask}& \multicolumn{4}{c}{Bounding Box Mask} \\
            \cmidrule(lr){2-5}
			\cmidrule(lr){6-9}
			&CS$\uparrow$ & IR$\uparrow$ &AS$\uparrow$ &CMMD$\downarrow$ &CS$\uparrow$ & IR$\uparrow$ &AS$\uparrow$ &CMMD$\downarrow$\\	
			\midrule
			BLD & 20.45 & -0.91 & 5.06 & 0.149 & 19.47 & -1.12 & 4.87 &0.085\\
            CNI & \underline{21.37} & -0.79 & 4.93 & 0.164 & \underline{20.97} & -0.95 & 4.83 & 0.095\\
            SDI & 20.23 & \underline{-0.60} & \underline{5.39} & 0.171 & 19.01 & \underline{-0.78} & \underline{5.31} & 0.109\\
            BN & 20.76 & -0.71 & 5.11 & \underline{0.143} & 20.13 & -0.87 & 5.04 & \underline{0.084}\\
            PP & \underline{21.37} & -0.66 & 5.10 & 0.152 & 20.92 & -0.80 & 5.12 & 0.104\\
			\midrule
			\textbf{Ours} & \textbf{23.49} & \textbf{-0.58} & \textbf{5.43} &  \textbf{0.136} & \textbf{23.46} & \textbf{-0.77} & \textbf{5.36} & \textbf{0.080}\\
			\bottomrule
		\end{tabular}
	\caption{Quantitative evaluation on Stylized-OpenImages dataset. }\vspace{-3mm}
	\label{tab:Quantitative_openimage}
\end{table}

\textit{2) Quantitative Evaluation:}  
Tab.~\ref{tab:Quantitative_mscoco} and Tab.~\ref{tab:Quantitative_openimage} present quantitative results on the Stylized-MSCOCO and Stylized-OpenImages datasets, demonstrating the effectiveness of HarmonPaint for inpainting stylized images. On the Stylized-MSCOCO dataset, we first evaluate inpainting with segmentation masks. Compared to PP, our method achieves a 0.74 improvement in CLIP Score (CS), highlighting the contributions of SAMS. SAMS primarily addresses structural fidelity, indirectly preserving the shape and layout. The IR metric is negative for all methods, likely due to the stylized test benchmarks being derived from style transfer techniques, which shift the images out of the original distribution. Despite this, IR serves as a valuable comparative reference. Our method achieves a 0.05 improvement in IR over SDI, demonstrating that the proposed MAKVS effectively captures and transfers style information, enhancing overall visual harmony.  
When using bounding box masks, which lack shape information about the inpainted content, our approach still performs robustly. SAMS enables the diffusion model to interpret the mask at the feature level, preserving structural fidelity and allowing our method to perform well even with minimal structural guidance. Similar trends are observed in Tab.~\ref{tab:Quantitative_openimage}, showing the effectiveness of HarmonPaint across diverse styles and masks.

\subsection{Ablation Study}
\textit{1) Components Analysis:}
To validate the contributions of each component in HarmonPaint, we first conduct a quantitative ablation study (Tab.~\ref{tab:ablation_study}). Incorporating the Self-Attention Masking Strategy (SAMS) improves the CLIP Score~\cite{hessel2021clipscore} by 0.92, demonstrating its effectiveness in enhancing structural fidelity within masked regions. The inclusion of $\mathcal{L}_s$ further increases the CLIP Score by 2.73, emphasizing its key role in aligning generation with the textual prompt. Meanwhile, the Mask-Adjusted Key-Value Strategy (MAKVS) contributes to stylistic coherence, leading to noticeable improvements in Image Reward (IR)~\cite{xu2024imagereward} by 0.2 and Aesthetic Score (AS)~\cite{schuhmann2022laion} by 0.1.

\begin{figure}
	\centering
	\includegraphics[width=\linewidth]{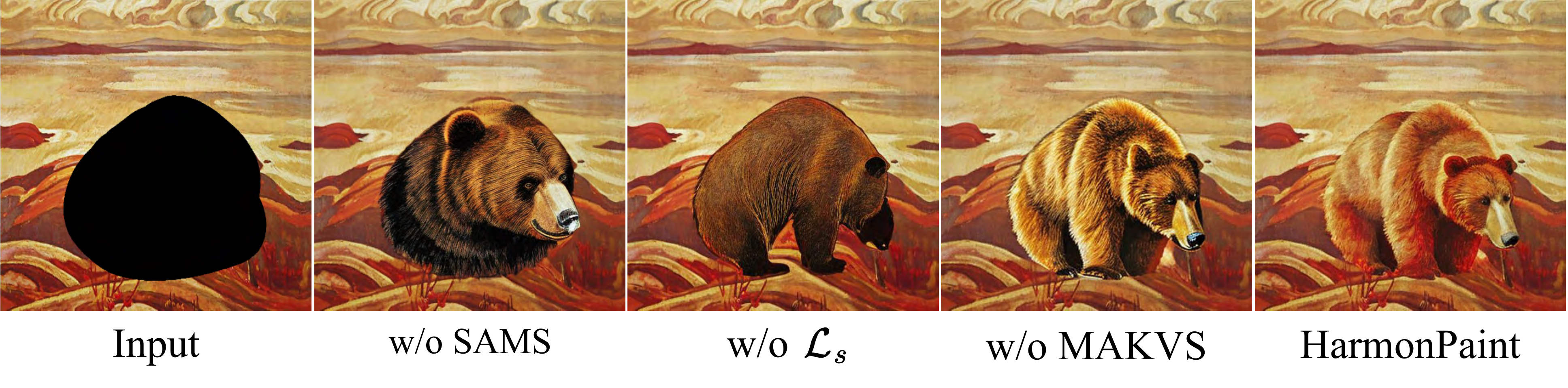}
	\vspace{-5mm}\caption{Qualitative ablation study on proposed components. }\vspace{-2mm}
	\label{fig:ablation study}
\end{figure}

We further corroborate these findings through a qualitative ablation study (Fig.~\ref{fig:ablation study}). Without SAMS, the model struggles to interpret masked regions, resulting in incomplete or spatially incoherent inpainting. Removing $\mathcal{L}_s$ causes attention to shift toward unmasked regions due to the lack of semantic guidance, thereby introducing prompt misalignment. Excluding MAKVS leads to stylistic fragmentation, where the inpainted content fails to integrate with its surrounding context. Together, these results underscore the importance and complementary nature of HarmonPaint’s core components in producing semantically aligned and visually harmonious images.

\begin{table}[t]
	\renewcommand\arraystretch{1.2}
	\centering	   
    {
		\begin{tabular}{l||c|c|c}
			\toprule
			Method &CS$\uparrow$ &IR$\uparrow$ &AS$\uparrow$ \\
			\midrule
			Baseline & 19.01 & -0.78 & 5.31 \\
            w/ SAMS & 19.93 & -0.88 & 5.21  \\
            w/ $\mathcal{L}_s$ & 21.74 & -0.89 & 5.28  \\
            w/o MAKVS &23.45 & -0.97 & 5.26  \\ 
			\midrule
			\textbf{HP (Ours)} & \textbf{23.46} & \textbf{-0.77} & \textbf{5.36} \\
			\bottomrule
		\end{tabular}}
	\caption{Quantitative ablation study results of different model variants.}
	\label{tab:ablation_study}
\end{table}

\begin{figure}
	\centering
	\includegraphics[width=\linewidth]{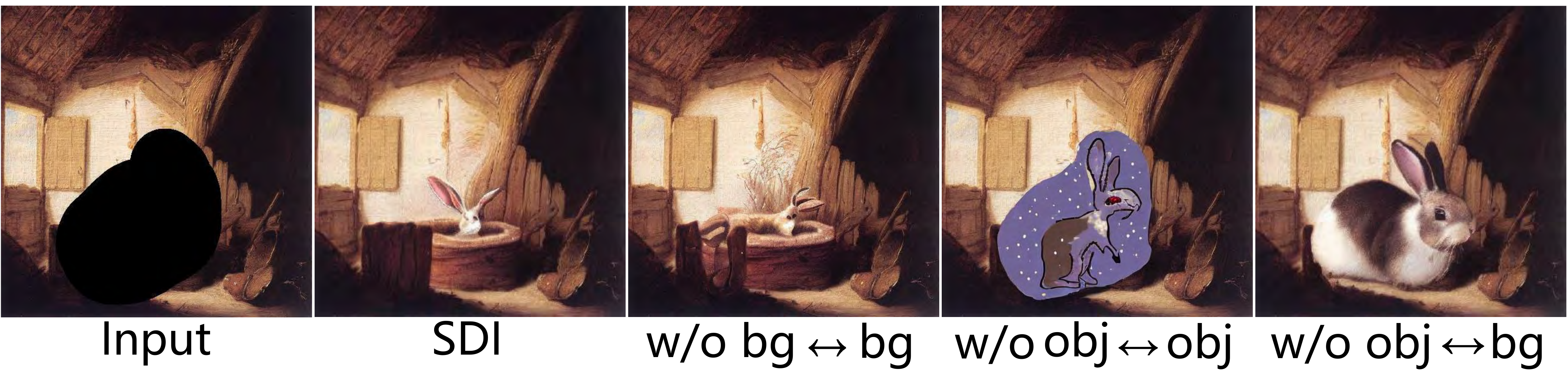}
	\vspace{-5mm}\caption{Experiment results of masking the background-background, the object-object, and object-background interactions.}\vspace{-2mm}
	\label{fig:spatial consistency}
\end{figure}

\textit{2) Effect of Mask Strategy in SAMS:}
To validate the structural fidelity of our Self-Attention Masking Strategy (SAMS), we partition the self-attention map into three regions, as analyzed in Section~\ref{section:contri1}, corresponding to interactions between object (obj) and background (bg). The results are shown in Fig.~\ref{fig:spatial consistency}. Masking bg-bg interactions shows little effect on background generation, as the concatenated input image and noise map already provide sufficient background priors. However, it does not benefit object generation. Conversely, masking obj-obj interactions significantly degrades object quality, as objects lack strong priors and depend more on object-text and object-object relations. Without obj-obj attention, the generated content maintains a coarse object shape (\textit{e.g.}, a blurry rabbit) but loses structural details. In comparison, our strategy, which masks only obj-bg interactions, effectively suppresses background interference while preserving object fidelity. Additional analysis on how SAMS shapes attention distributions is provided in the supplementary materials.
\begin{figure}
	\centering
	\includegraphics[width=\linewidth]{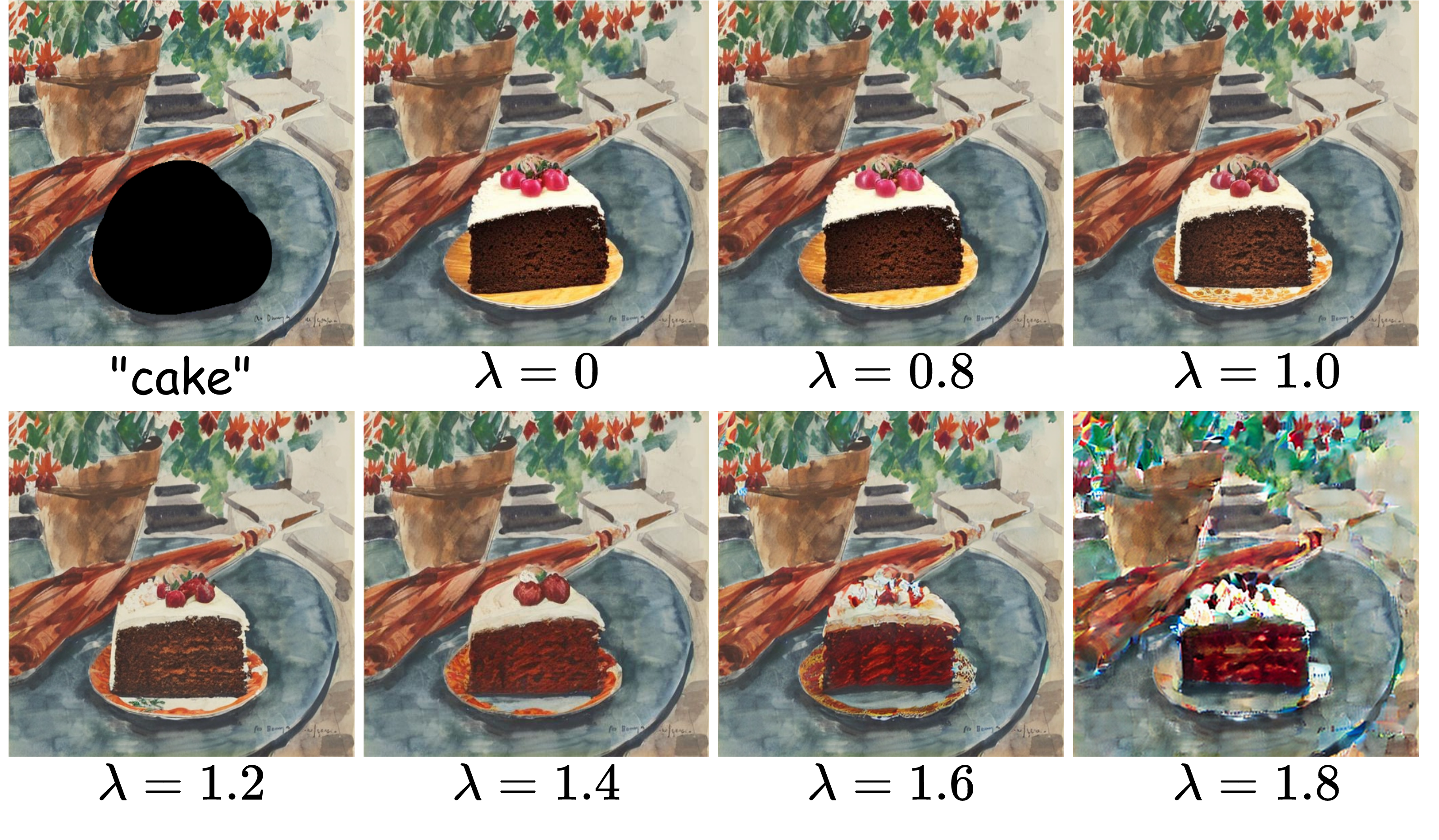}
	\caption{Visualization of the effects of $\lambda$. As $\lambda$ increases, the stylistic harmony of the image improves. However, when $\lambda$ exceeds 1.4, the quality of the inpainted content begins to noticeably decline. }\vspace{-3mm}
	\label{fig:lambda}
\end{figure}

\begin{figure}
	\centering
	\includegraphics[width=\linewidth]{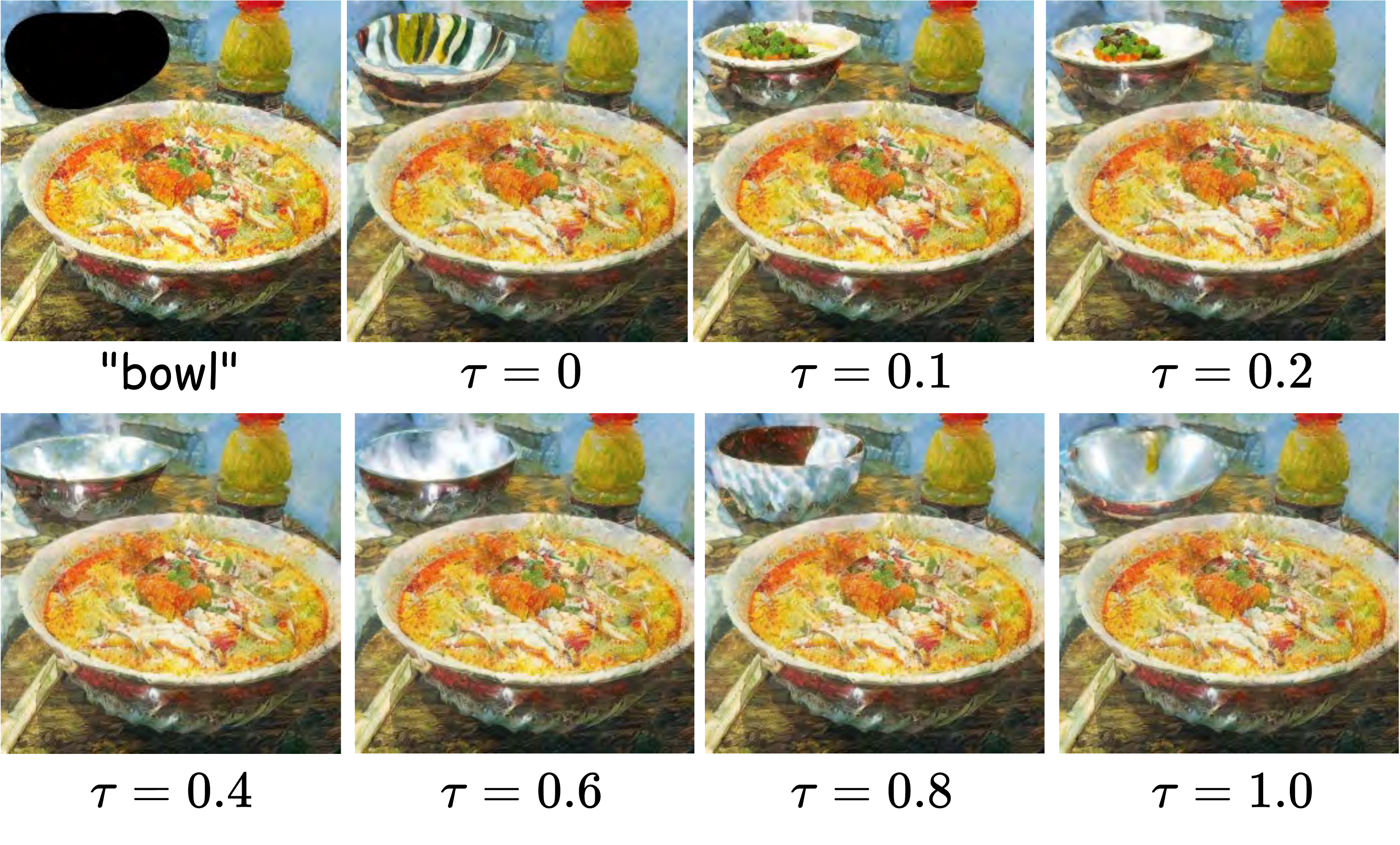}
	\caption{Visualization of the effects of $\tau$. As $\tau$ increases, the structural fidelity of the image improves. However, when $\lambda$ exceeds 0.8, the quality of the inpainted content begins to noticeably decline.  }\vspace{-4mm}
	\label{fig:tau}
\end{figure}
\textit{3) Hyperparameter Analysis:}
To further examine the impact of $\lambda$ on stylistic harmony, we show inpainting results for varying $\lambda$ values in Fig.~\ref{fig:lambda}. With $\lambda$ set to 0, indicating that MAKVS is inactive, the ``cake'' in the masked region appears close to a real-world style. As $\lambda$ increases, the style of the ``cake'' progressively aligns with that of the unmasked region, achieving visual harmony at $\lambda = 1.4$. However, increasing $\lambda$ beyond this point causes the self-attention layer to overemphasize style information, which interferes with content generation.

We also investigate the effect of the parameter $\tau$ on structural fidelity by presenting inpainting results under different $\tau$ values in Fig~\ref{fig:tau}. Our experiments show that setting $\tau=0$ (hard mask) leads to noticeable boundary artifacts. In contrast, values of $\tau$ within the range 0.1 to 0.6 strike an optimal balance, yielding high-quality, well-controlled inpainting results. However, when $\tau \geq 0.8$, the inpainting often fails, likely due to the excessive softening of the mask, which blurs the distinction between masked and unmasked regions.

\subsection{General Inpainting}  
\begin{figure}
	\centering
	\includegraphics[width=\linewidth]{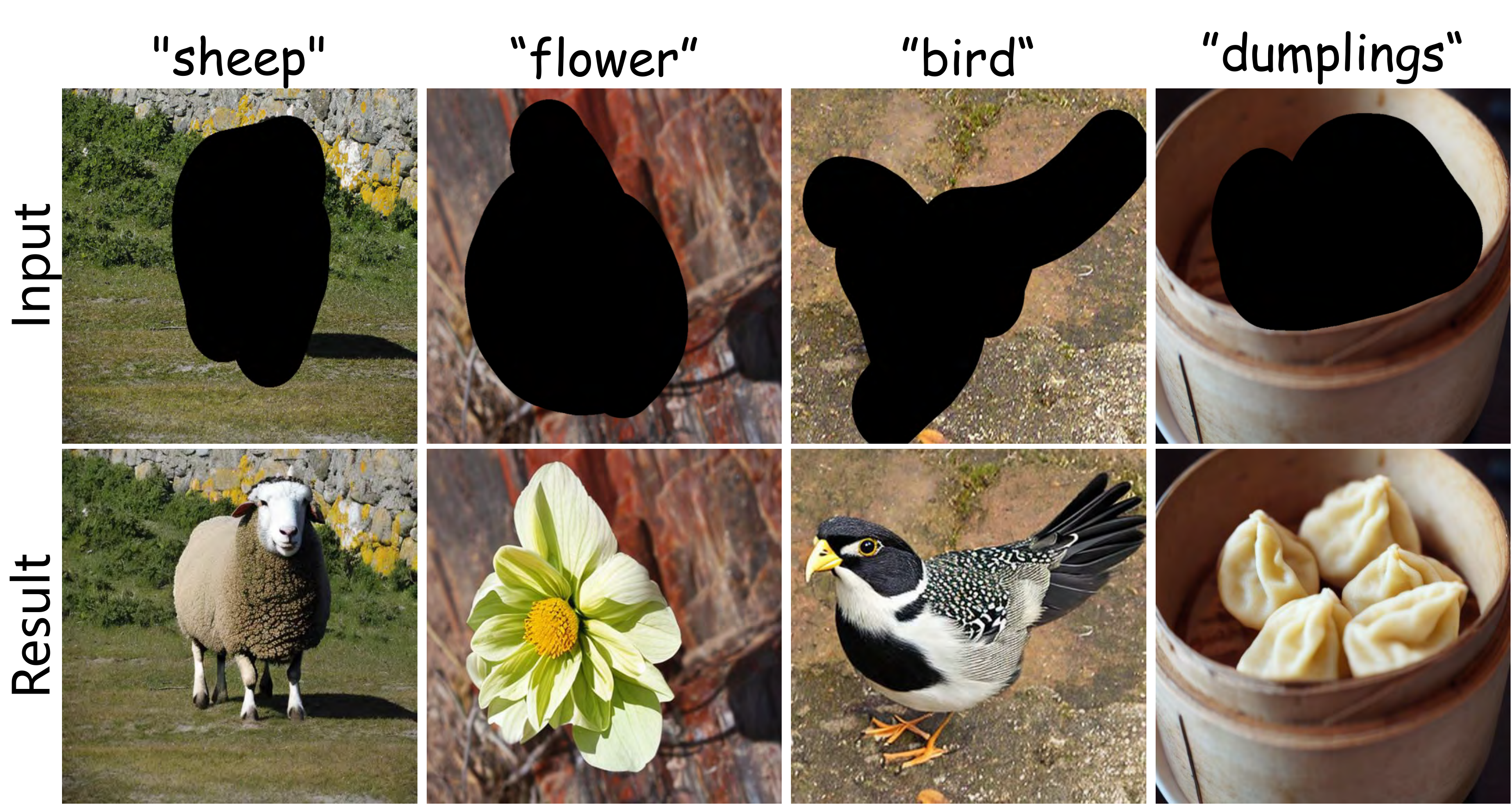}
	\caption{The results of our method applied to general inpainting. }\vspace{-1mm}
	\label{fig:general}
\end{figure}

Although our method is primarily designed for stylized image inpainting, it also performs effectively on general inpainting tasks, as shown in Fig.~\ref{fig:general}. Unlike stylized images, natural images lack a distinct artistic style, yet achieving visual harmony still requires careful attention to lighting and color consistency. To address this, we set $\lambda$ to 0.8, allowing the diffusion model to capture subtle visual attributes while minimizing stylistic interference from the unmasked region. The results align with the text descriptions while maintaining a realistic and cohesive appearance, demonstrating the versatility of our approach. Notably, while prior methods struggle to adapt to both stylized and natural image inpainting, our method excels in both scenarios without requiring additional modifications.

\begin{figure}
	\centering
	\includegraphics[width=0.9\linewidth]{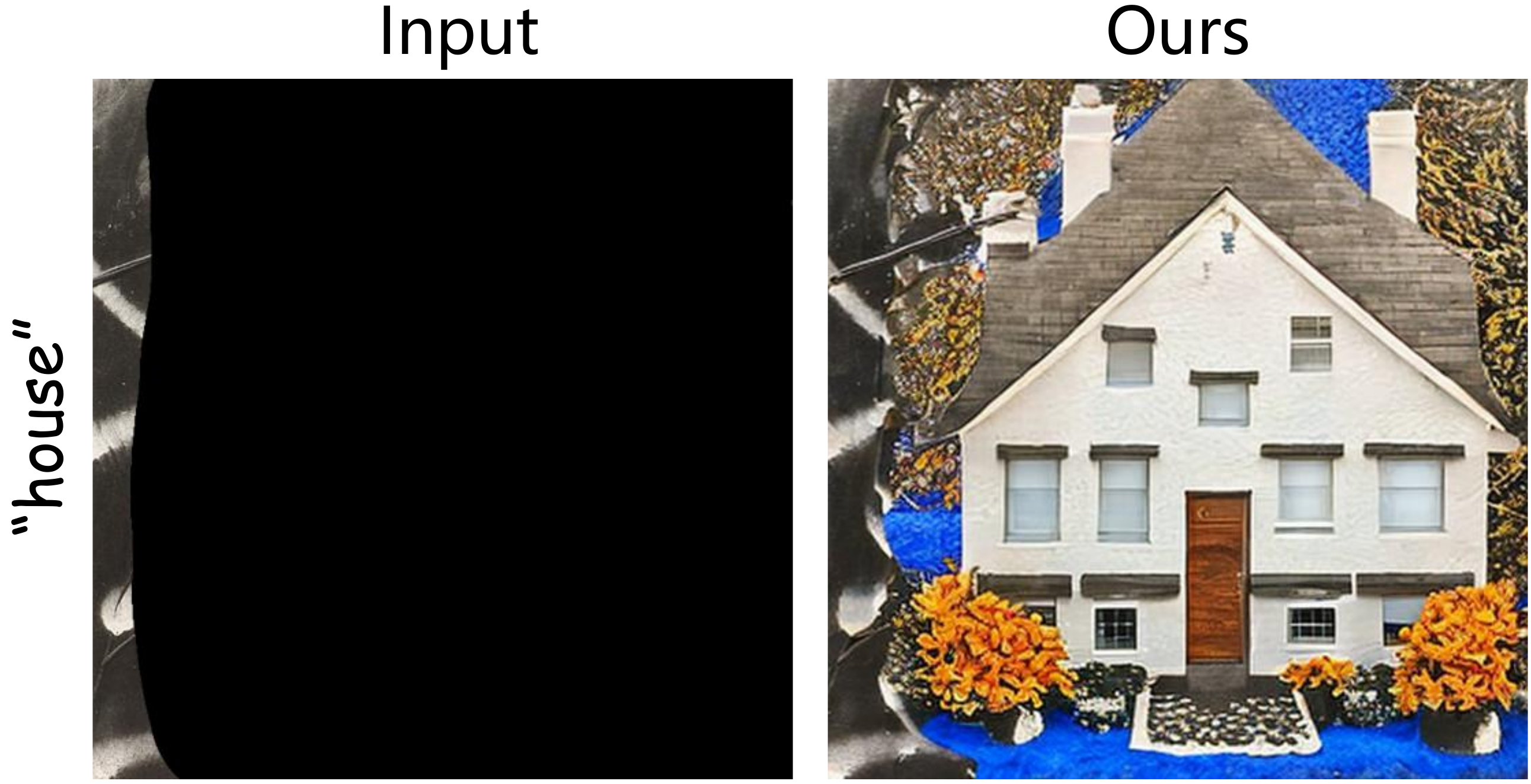}
	\caption{When the image is more than 90\% missing, especially when only one edge remains, although we can use text to complete the masked content, it is difficult to achieve Harmonized Inpainting due to the insufficient style information provided outside the mask.}\vspace{-4mm}
	\label{fig:failure cases}
\end{figure}

%% file: sec/5_conclusion.tex
\section{Conclusions, Limitations, and Future Work}
\label{sec:con}
In this paper, we propose HarmonPaint, a diffusion-based inpainting framework capable of producing harmonious and structurally coherent results across diverse styles without requiring additional training. HarmonPaint effectively preserves structural fidelity and ensures stylistic consistency between inpainted and unmasked regions. Extensive experiments demonstrate its robustness under various mask types and its applicability to natural images.

While our method performs well for small to medium missing regions, its effectiveness degrades when the masked area exceeds 90\% of the image, as shown in Fig.~\ref{fig:failure cases}. This limitation arises from the reliance on the remaining unmasked content as the primary source of style cues, which becomes insufficient in extreme cases. In future work, we plan to explore strategies to infer stylistic information directly from text descriptions or to incorporate external references to improve performance in scenarios with large missing regions.

%% file: supp.tex
\clearpage
\appendices
\section{User Study}  
We conduct a user study to evaluate the performance of our method against competing approaches. A total of 20 incomplete images are randomly selected, and their inpainting results using our method and other methods are compiled into a questionnaire with two evaluation criteria: Q1) ``Which image better aligns with the given text description?'' and Q2) ``Which image exhibits stronger style consistency and visual harmony?'' The study is completed by 40 participants independently, without prior knowledge of the methods used. Scores are computed as the proportion of times each method is selected for the two questions, with results summarized in Tab.~\ref{tab:user_study}.  
\begin{figure*}
	\centering
	\includegraphics[width=0.9\linewidth]{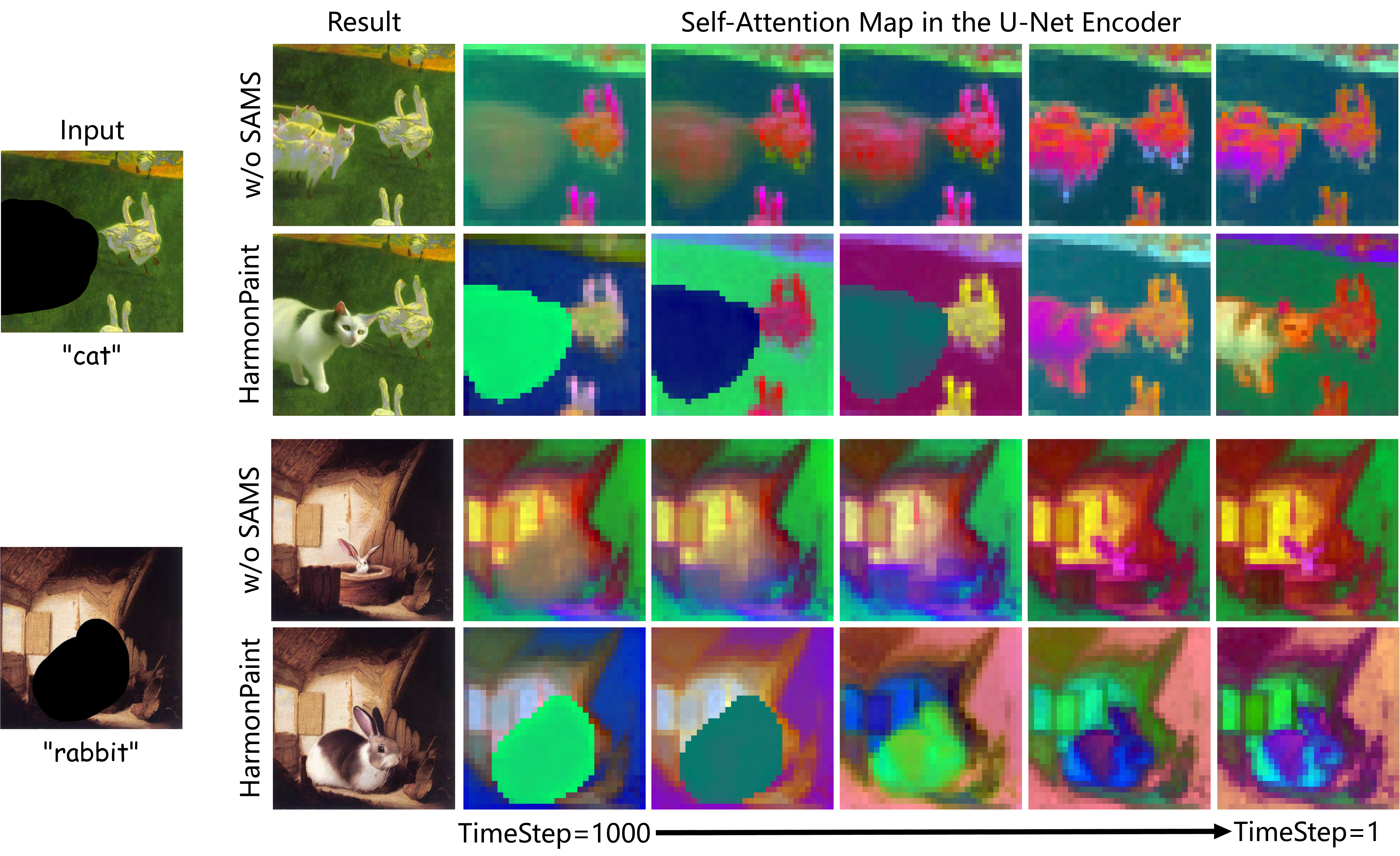}
	\caption{Additional PCA visualizations of self-attention maps show that as timesteps decrease, our method effectively generates coherent shapes and contours within masked regions.
	}
	\label{fig:SAMS}
\end{figure*}
Our method achieves the highest scores in both questions, demonstrating its ability to ensure strong text alignment and structural fidelity simultaneously. While methods such as Brushnet and PowerPaint performed well in Q1, they scored lower in Q2 due to insufficient consideration of stylistic harmony. HarmonPaint addresses this limitation using the Mask-Adjusted Key-Value Strategy (MAKVS) to transfer style information from unmasked regions into the masked areas, ensuring visual coherence. Additionally, the introduced $\mathcal L_s$ focuses the cross-attention on the masked region, resulting in superior text alignment and securing the highest user preference in Q1.  

\begin{table}[h]
	\renewcommand\arraystretch{1.2}
	\centering	
    \setlength\tabcolsep{1mm}	
    {
		\begin{tabular}{l||c|c}
			\toprule
			Method &Q1 &Q2 \\
			\midrule
			Blend Latent Diffusion~\cite{avrahami2023blended} & 4.000\% & 2.875\% \\
            ControlNet-Inpainting~\cite{zhang2023adding} & 9.125\% & 7.625\%  \\
            Stable Diffusion-Inpainting~\cite{rombach2022high} &8.500\% & 6.625\%  \\ 
            BrushNet~\cite{ju2024brushnet} & 11.875\% & 10.250\%  \\
            PowerPaint~\cite{zhuang2023task} & 13.250\% & 12.125\%  \\	
			\midrule
			\textbf{HP (Ours)} & \textbf{53.250\%} & \textbf{60.500\%} \\
			\bottomrule
		\end{tabular}}
	\caption{User study results. Our method achieves the highest percentage, reflecting stronger alignment with user preferences in terms of text description accuracy and stylistic harmony.
}
	\label{tab:user_study}
\end{table}

\begin{figure*}
	\centering
	\includegraphics[width=0.9\linewidth]{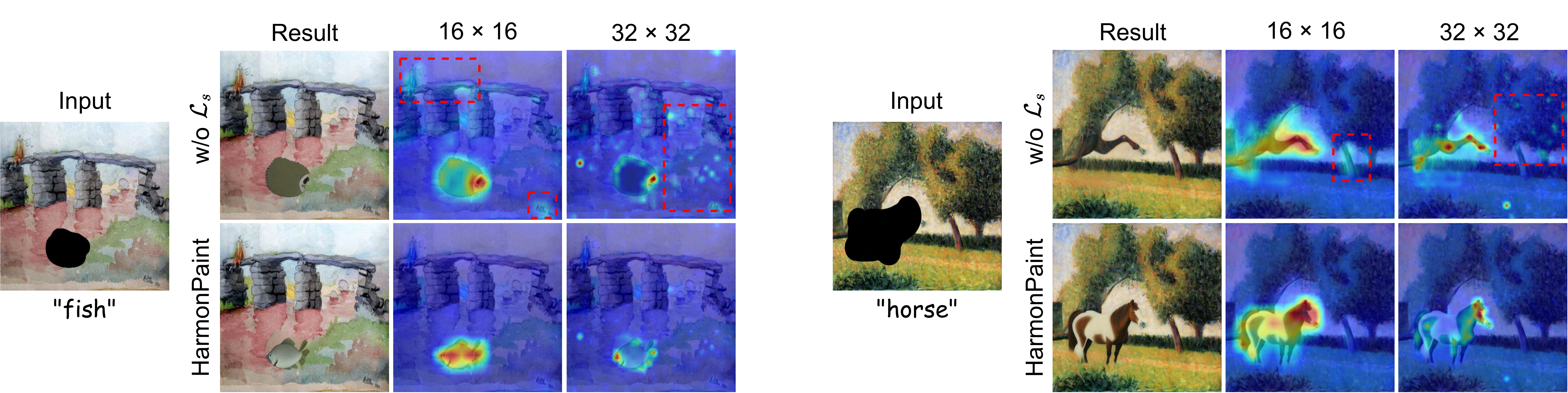}
	\caption{Additional cross-attention map visualizations at resolutions 16 and 32. Red boxes indicate instances where the text fails to focus attention on the masked region.}
	\label{fig:att_steering}
\end{figure*}

\section{Additional Analysis}
\noindent\textbf{The Role of the Self-Attention Masking Strategy (SAMS).}  
To validate the effectiveness of SAMS, we extract self-attention maps from the U-Net encoder and visualize them using PCA, as shown in Fig.~\ref{fig:SAMS}. Without SAMS, the masked and unmasked regions exhibit similar colors, indicating that the U-Net struggles to differentiate their features. This results in the inpainted content within the masked area being influenced by features from unmasked regions, causing spatial inconsistencies as the timestep decreases. In contrast, our method incorporates SAMS, enabling the U-Net to clearly distinguish between features inside and outside the mask. This distinction is reflected in the self-attention maps, where the masked region consistently displays distinct colors from the surrounding areas throughout the denoising process, ensuring spatial coherence.  

\begin{figure}
	\centering
	\includegraphics[width=\linewidth]{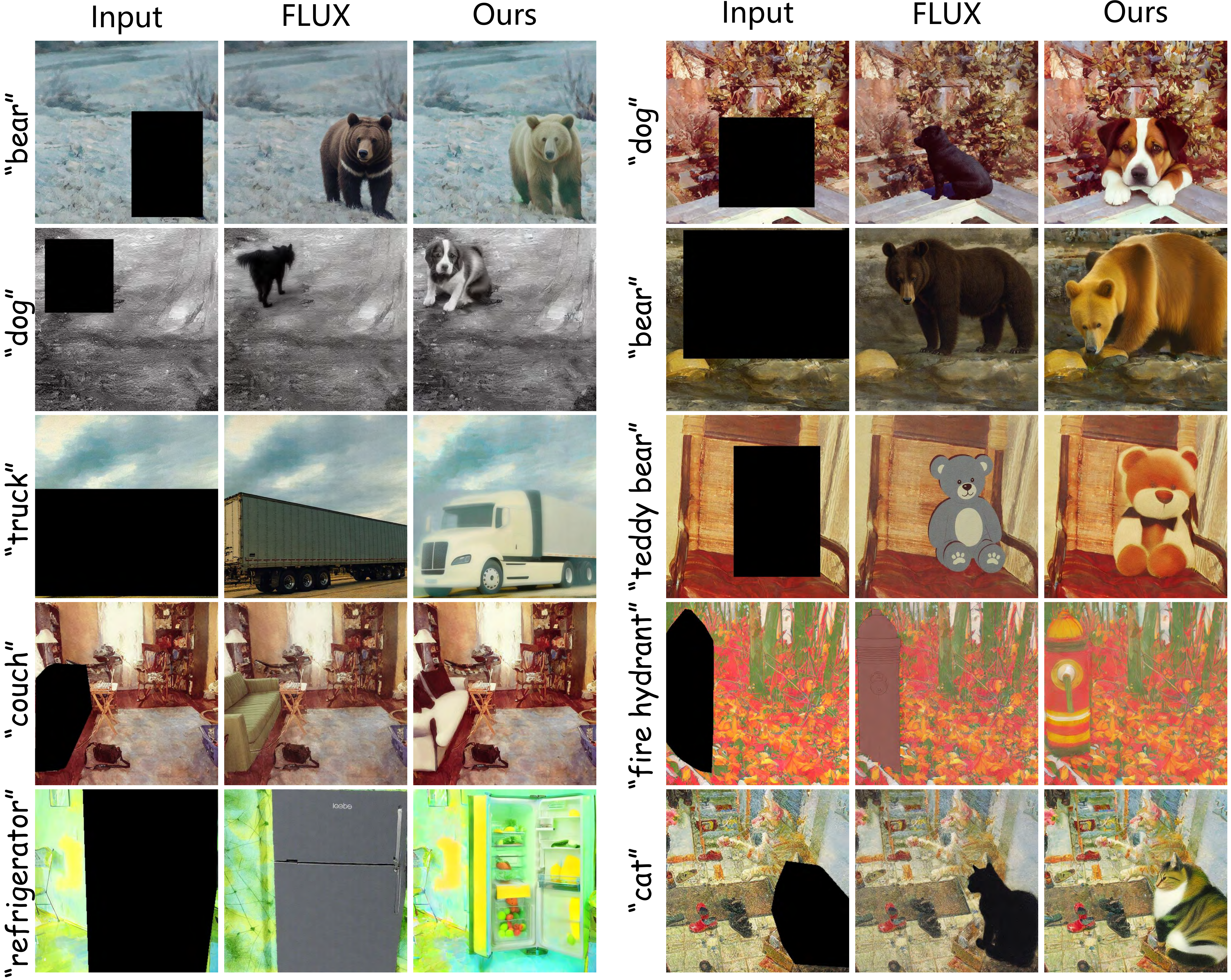}
	\caption{Qualitative comparison with FLUX.}
	\label{fig:flux}
\end{figure}

\noindent\textbf{The Role of $\mathcal{L}_{s}$.}  
Fig.~\ref{fig:att_steering} illustrates the impact of $\mathcal{L}_{s}$ with additional comparisons before and after its application. Without $\mathcal{L}_{s}$, the cross-attention is dispersed across multiple patches of the image, failing to concentrate on the target area within the mask. This leads to inpainting results that are poorly aligned with the text prompt. By incorporating $\mathcal{L}_{s}$, our method effectively refines the noise map, ensuring that attention is focused on the masked region. This refinement enhances the alignment between the inpainted results and the text description, improving semantic consistency.  

\section{Comparison with FLUX}
To validate the effectiveness of our method, we conduct comparative experiments with FLUX~\cite{flux2024}, a state-of-the-art model known for its strong inpainting capabilities, using both qualitative and quantitative evaluations. As shown in Fig.~\ref{fig:flux}, our method demonstrates superior stylistic harmony compared to FLUX. When given the user prompt ``cat'', FLUX generates a black cat that, while structurally accurate, exhibits noticeable chromatic dissonance against the background. In contrast, our method leverages MAKVS to generate a cat that not only preserves structural accuracy but also aligns stylistically with the surrounding content, resulting in more visually coherent and balanced outputs.

For quantitative analysis, we evaluate both methods on the Stylized MSCOCO dataset using three metrics: CLIP Score (CS), Image Reward (IR), and Aesthetic Score (AS). As shown in Table~\ref{tab:flux}, our method outperforms FLUX by margins of 2.14, 0.23, and 1.23 on the segmentation mask, respectively. These quantitative results, together with the qualitative findings, validate the technical advancement and practical effectiveness of our method in image inpainting tasks.

\begin{table}[h]
	\renewcommand\arraystretch{1.2}
	\centering
     \setlength\tabcolsep{1mm}
		\begin{tabular}{l||c|c|c|c|c|c}
			\toprule
			\multirow{2}*{Method}  
            & \multicolumn{3}{c|}{Segmentation Mask}& \multicolumn{3}{c}{Bounding Box Mask} \\
            \cmidrule(lr){2-4}
			\cmidrule(lr){5-7}
			&CS$\uparrow$ & IR$\uparrow$ &AS$\uparrow$ &CS$\uparrow$ & IR$\uparrow$ &AS$\uparrow$ \\	
			\midrule
			FLUX & 26.72 & -0.79 & 5.32 & 26.81 & -0.73 & 5.33 \\
			\midrule
			\textbf{Ours} & \textbf{28.86} & \textbf{-0.56} & \textbf{6.55} & \textbf{28.27} & \textbf{-0.46} & \textbf{5.96}\\
			\bottomrule
		\end{tabular}	
    \captionsetup{width=3\linewidth,justification=centering}
    \caption{Quantitative comparison with FLUX.}
    \vspace{-4mm}
	\label{tab:flux}
    \end{table}
    
\begin{figure}
	\centering
	\includegraphics[width=0.9\linewidth]{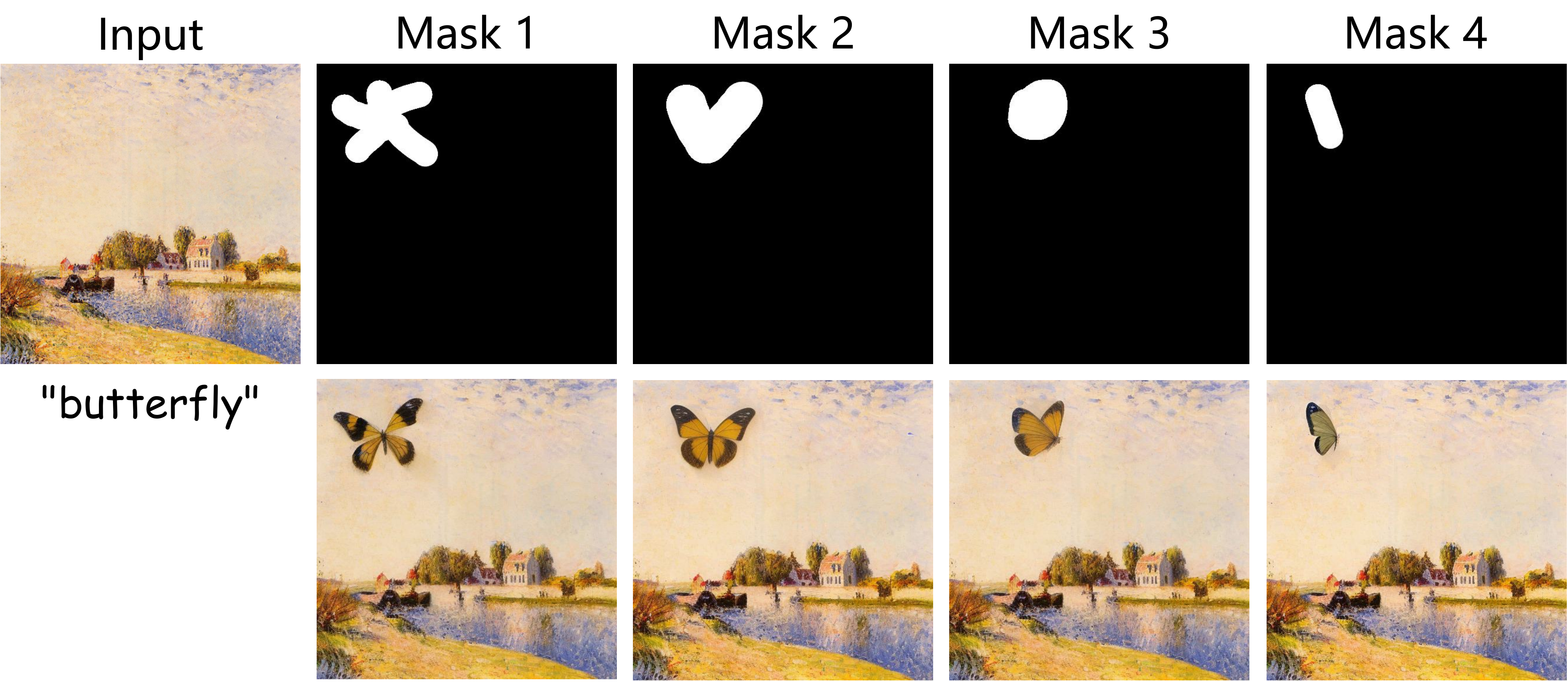}
	\caption{Our method demonstrates strong adaptability to various mask types, generating results that align with the mask shape and remain consistent with the text description.}\vspace{-4mm}
	\label{fig:mask-setting1}
\end{figure}

\section{Mask Sensitivity}  
In real-world applications, user-provided masks often vary significantly in form and precision, usually offering only an approximate outline of the target's location and shape. To evaluate the robustness of our method to such mask variability, we designed experiments that simulate practical scenarios users might encounter.

We manually created different mask shapes on the same image and performed inpainting with the prompt ``butterfly,'' as shown in Fig.~\ref{fig:mask-setting1}. In the first and second columns, rough shapes approximating a ``butterfly'' were used as masks, and HarmonPaint successfully adapts to these forms, capturing relevant details such as the butterfly’s wings. To account for cases where users may provide random mask shapes with no direct correlation to the prompt, we tested simpler masks in the third and fourth columns. Our method still generates plausible shapes within the masked areas. This is due to our SAMS, which enables the diffusion model to effectively interpret the spatial relationships between masked and unmasked regions, resulting in coherent shapes and contours within the masked region.

\begin{figure}
	\centering
	\includegraphics[width=0.9\linewidth]{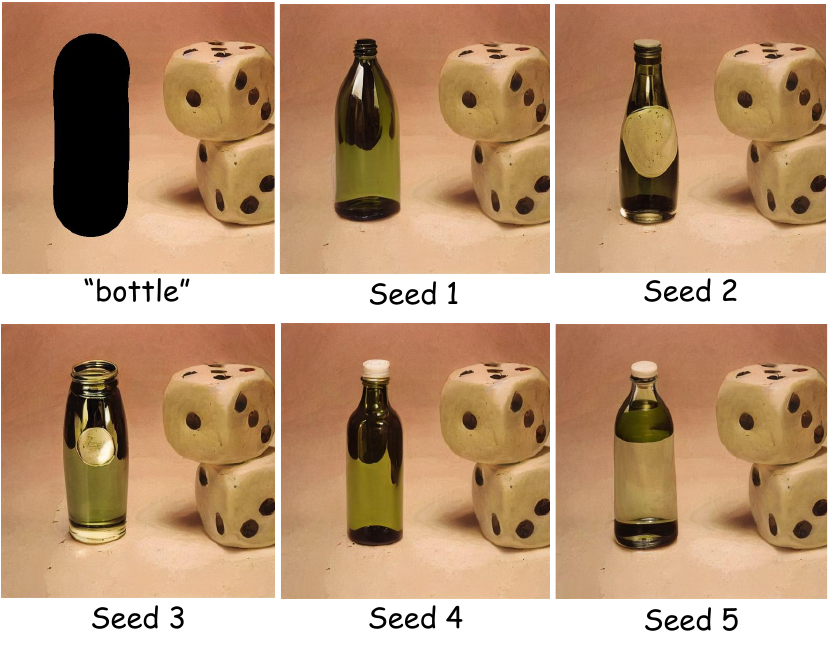}
	\caption{Visualization of model output variability under same Prompts.}
	\label{fig:Model diversity}
\end{figure}

\section{Output Diversity}
As illustrated in Fig.~\ref{fig:Model diversity}, our model demonstrates pronounced semantic and visual diversity across generations conditioned on the same input prompt. Although the incorporation of SAMS and the loss $\mathcal L_s$ improves structural fidelity, these mechanisms primarily operate at a local level without enforcing rigid global constraints. Consequently, the model retains the generative flexibility inherent to the baseline Stable Diffusion Inpainting, producing outputs with substantial variation in both structure and style. The localized nature of attention-based guidance allows the model to maintain creative variability while ensuring alignment with essential structural cues. This balance between consistency and expressive diversity is particularly advantageous in tasks that demand both structural coherence and rich variation.

\section{Partial Object Inpainting}
\begin{figure}
	\centering
	\includegraphics[width=0.9\linewidth]{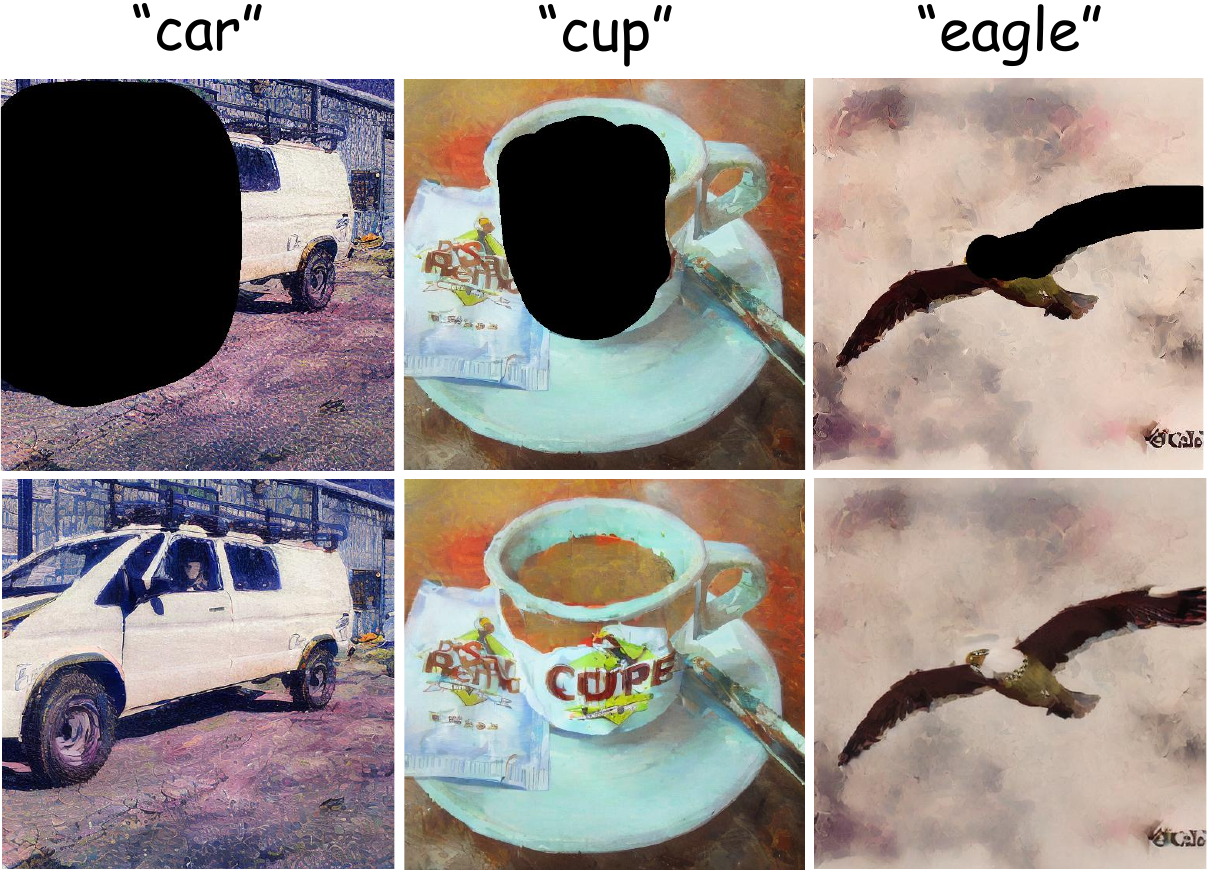}
	\caption{Visualization of partial inpainting.}\vspace{-4mm}
	\label{fig:Partial Inpainting}
\end{figure}

As shown in Fig.~\ref{fig:Partial Inpainting}, our model extends beyond full-object completion in stylized images to effectively handle partial object inpainting. The proposed SAMS module provides adaptive guidance that enhances generation within masked regions without introducing unnecessary rigidity. As a result, the inherent local completion capability of Stable Diffusion Inpainting remains intact. Moreover, by leveraging style cues from unmasked areas, our approach ensures stylistic coherence and seamless integration with the surrounding content.

\section{Additional Visual Results}  
\begin{figure*}
	\centering
	\includegraphics[width=\linewidth]{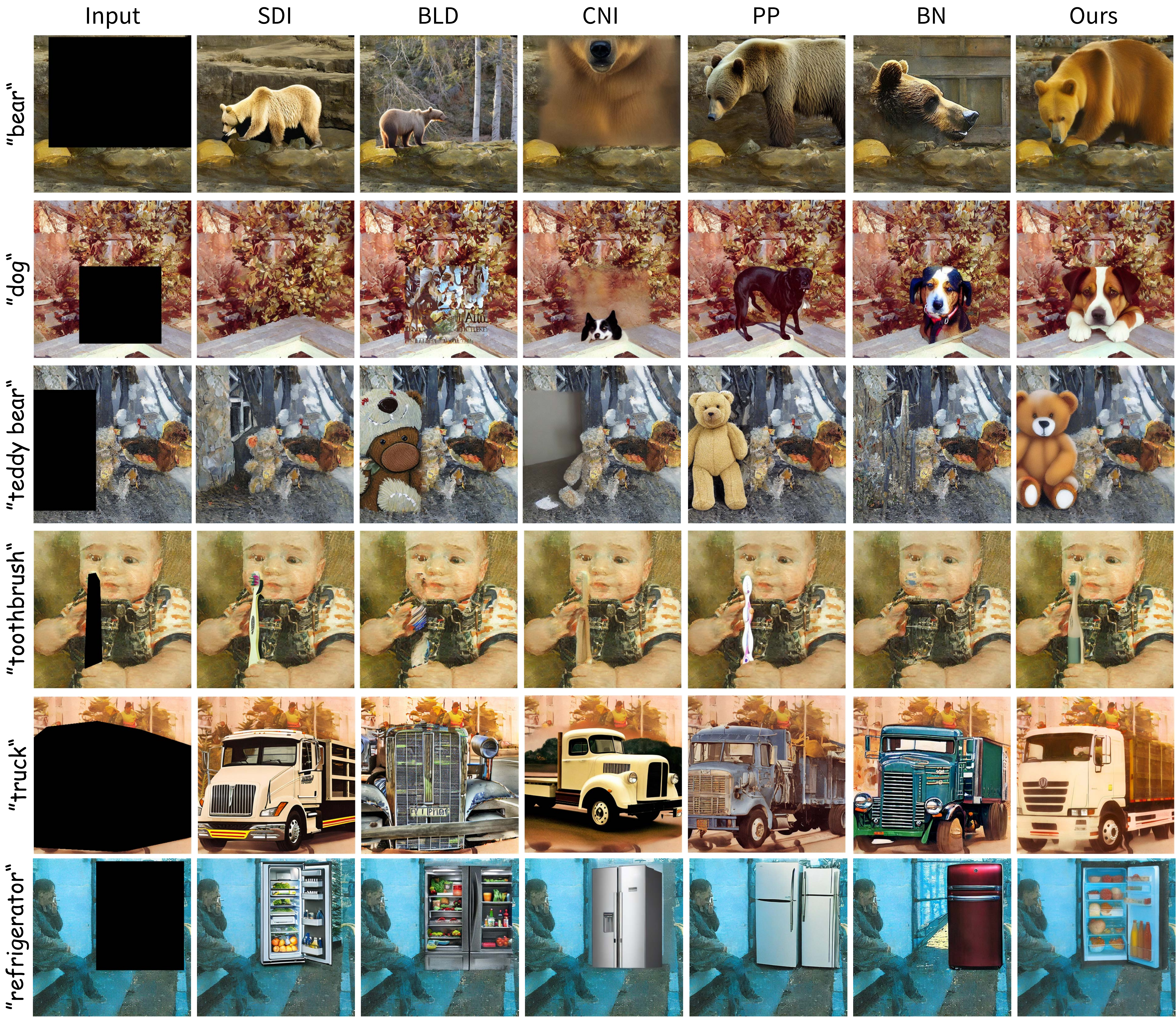}
	\caption{Additional qualitative comparisons on the Stylized-COCO dataset.}
	\label{fig:supp_mscoco}
\end{figure*}

\begin{figure*}
	\centering
	\includegraphics[width=\linewidth]{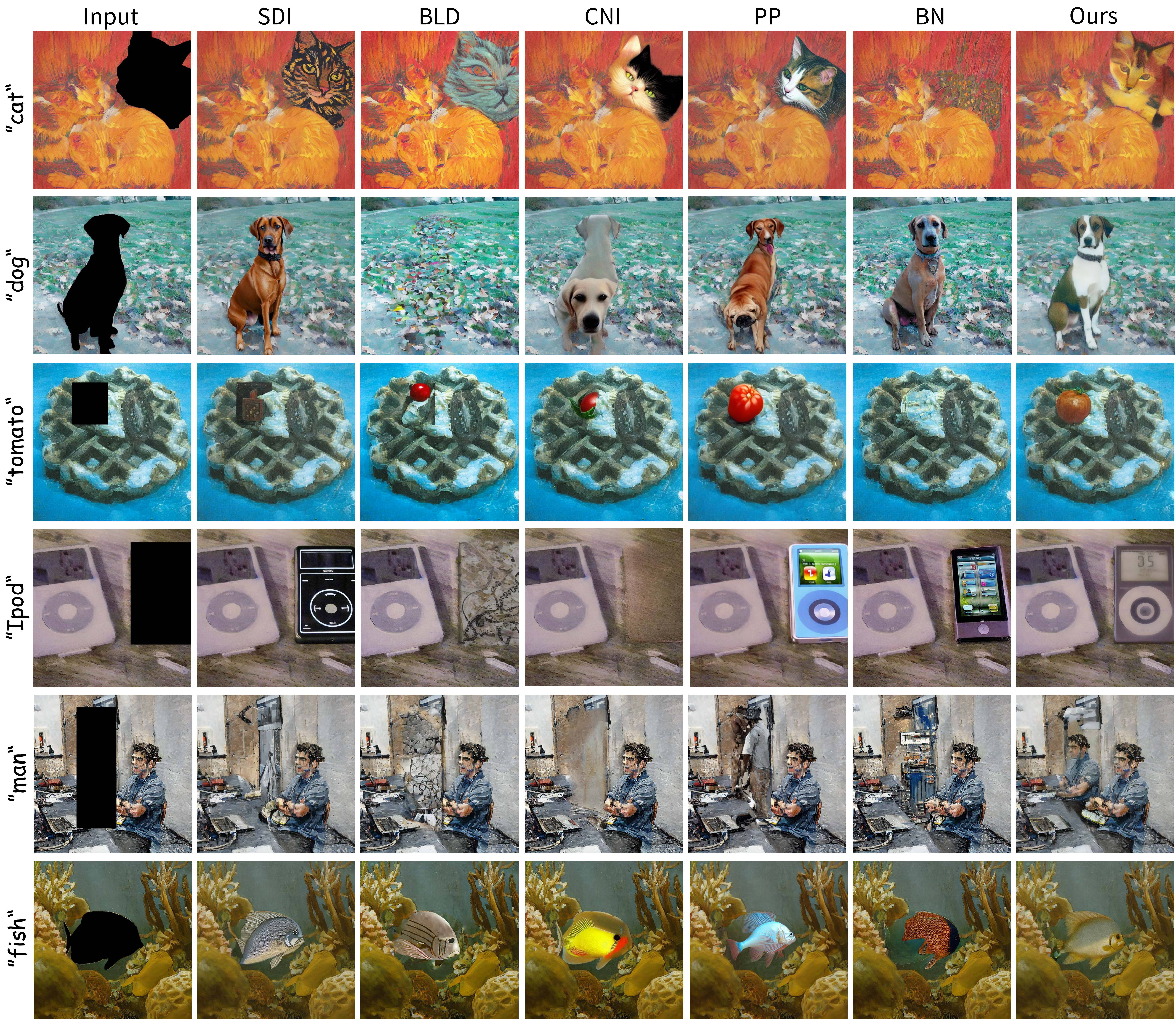}
	\caption{Additional qualitative comparisons on the Stylized-OpenImages dataset. }
	\label{fig:supp_openimages}
\end{figure*}

We provide additional qualitative comparisons with other competitors in Fig.~\ref{fig:supp_mscoco} and Fig.~\ref{fig:supp_openimages}, further demonstrating the superiority of our approach. A demo is also included in the supplementary materials to offer a more detailed visual validation of our method's effectiveness.

